\pdfoutput=1
\PassOptionsToPackage{svgnames,dvipsnames}{xcolor}

\documentclass[11pt]{article}

\usepackage[final]{acl}

\usepackage{times}
\usepackage{latexsym}

\usepackage{booktabs}
\usepackage{multirow}

\usepackage{float} 

\usepackage{tabularx}

\usepackage{longtable}
\usepackage{parcolumns}
\usepackage{booktabs}

\usepackage[T1]{fontenc}
\usepackage[utf8]{inputenc}

\usepackage{subcaption}

\usepackage{microtype}

\usepackage{amsfonts}

\usepackage{colortbl}

\usepackage{placeins}

\usepackage{arydshln}

\usepackage{inconsolata}

\usepackage{graphicx}

\usepackage{tikz} 
\usepackage[most]{tcolorbox}  
\usepackage{soul}
\usepackage{enumitem}  

\definecolor{promptbox}{RGB}{248, 249, 250}
\definecolor{optional}{RGB}{70, 130, 180}  
\definecolor{parameter}{RGB}{222, 83, 138}
\definecolor{optionalbg}{RGB}{240, 248, 255}  
\definecolor{paramheading}{RGB}{90, 90, 90}  

\newcommand{\param}[1]{{\color{parameter}\texttt{<#1>}}}

\newsavebox{\optionalbox}
\newcommand{\optional}[1]{%
  \sbox{\optionalbox}{%
    \begin{minipage}{\linewidth}%
      \begin{tcolorbox}[
        boxrule=0.5pt,
        colback=optionalbg,
        colframe=optional,
        left=8pt,
        right=8pt,
        top=5pt,
        bottom=5pt,
        arc=2pt,
        boxsep=0pt,
        enhanced,
        breakable,
        before skip=6pt,
        after skip=6pt,
        overlay={
          \node[fill=optional!30, text=optional, font=\small, inner sep=3pt, 
                anchor=south east, rounded corners=3pt]
               at ([xshift=-2pt, yshift=2pt]frame.south east) {Optional};
        }
      ]%
        {\color{optional}#1}%
      \end{tcolorbox}%
    \end{minipage}%
  }%
  \usebox{\optionalbox}%
}

\newcommand{\optionalinline}[1]{%
  \begingroup
  \sethlcolor{optionalbg}%
  {\color{optional}\small\hl{#1}}%
  \endgroup
}

\newtcolorbox{promptbox}[2][]{
  colback=promptbox,
  colframe=black!80,
  arc=2mm,
  boxrule=1pt,
  left=10pt,
  right=10pt,
  top=8pt,
  bottom=8pt,
  breakable,
  before skip=12pt,
  after skip=12pt,
  fonttitle=\bfseries,
  title=#2,
  #1 
}

\newcommand{\promptsep}{%
  \vspace{1em}{\color{gray!30}\hrule height 0.5pt}\vspace{1em}%
}

\newcommand{\paramheading}{{\textbf{Parameters}}}

\newenvironment{paramlist}{%
  \begin{itemize}[itemsep=2pt,parsep=0pt]%
}{\end{itemize}}

\title{The AI Co-Ethnographer:\\How Far Can Automation Take Qualitative Research?}

\author{\textbf{Fabian Retkowski}$^{1}$, Andreas Sudmann$^{2}$, \textbf{Alexander Waibel}$^{1,3}$\\
  $^1$Karlsruhe Institute of Technology, Germany\\
  $^2$University of Bonn, Germany\\
  $^3$Carnegie Mellon University, USA\\
  \texttt{\{retkowski,waibel\}@kit.edu} / \texttt{asudmann@uni-bonn.de}
}

\begin{document}
\maketitle
\begin{abstract}
Qualitative research often involves labor-intensive processes that are difficult to scale while preserving analytical depth. This paper introduces The AI Co-Ethnographer (AICoE), a novel end-to-end pipeline developed for qualitative research and designed to move beyond the limitations of simply automating code assignments, offering a more integrated approach. AICoE organizes the entire process, encompassing open coding, code consolidation, code application, and even pattern discovery, leading to a comprehensive analysis of qualitative data.
\end{abstract}

\section{Introduction}

Qualitative data analysis is a crucial research approach in the humanities, cultural studies, and social sciences, focusing on the synchronic and diachronic analysis and interpretation of non-numerical data such as texts, images, or audio files to gain insights into complex social phenomena, cultural expressions, and individual experiences \cite{creswell_qualitative_2017, denzin_sage_2023}. 
Coding is central to this process, structuring and interpreting research materials such as interviews, field notes, or group discussions by systematically assigning analytically relevant concepts to text segments or other data forms \cite{holton_coding_2007,bernard_research_2011,harding_qualitative_2013, bernard_analyzing_2016}.

Although coding offers a formalized structure for data analysis, its application remains context-specific and flexible, adapting to the nuances of the research question and subject matter \cite{elliott_thinking_2018}. In many contexts, specifically in ethnographic approaches, coding is inherently iterative and closely tied to an ongoing process of collecting and reflecting on data. 
Codes evolve dynamically through an iterative process where they are merged, adjusted, added, or replaced as researchers engage with the data, identify patterns, and refine their conceptual understanding. This process may involve open or axial coding, deductively or inductively, quantitatively or qualitatively, and can be centered on interpretation or description. \cite{ritchie_qualitative_2014,creswell_30_2015,saldana_coding_2015}.

However, manual coding faces significant limitations. Scalability remains a critical challenge when researchers encounter larger datasets that require extensive time and resources to code effectively \cite{miles_qualitative_2019}. It also increases the risk of intra- and intercoder unreliability, just to mention a few typical challenges.
These constraints have spurred interdisciplinary efforts to automate the coding process over the past decade. Automated speech recognition (ASR) has emerged as a significant enabler in this landscape, allowing researchers to efficiently transcribe large volumes of interview data and prepare them for further analysis and processing \cite{nguyen_super-human_2021}. Related qualitative data processing tasks such as text summarization \cite{hori_automatic_2002,retkowski_zero-shot_2024,zhang_benchmarking_2024}, question answering \cite{singhal_toward_2025}, and topic segmentation \cite{zechner_diasumm_2000,retkowski_text_2024} have similarly benefited from computational advancements, providing researchers with tools to condense information and identify thematic boundaries.

Recently, large language models (LLMs) have demonstrated new epistemic capabilities to annotate research data, yet with certain limitations, such as understanding the broader context of codes \cite{tuschling_chatgpt_2023,fischer_exploring_2024,rasheed_can_2024,ziems_can_2024}. In parallel, the concept of Agentic LLMs has emerged, designed to operate autonomously with goal-directed behaviors \cite{xi_rise_2023}. For example, the \textit{AI Scientist} \cite{lu_ai_2024} showcased an end-to-end automated workflow for writing scientific papers, from hypothesis generation, experimental design and manuscript drafting. This work illustrates the potential for autonomous agents to manage complex, multi-stage research processes. Inspired by these advances, our approach seeks to explore similar automation in the domain of qualitative research, also as an alternative to AI-assisted data analysis with proprietary systems like MaxQDA.

\begin{figure*}[th!]
\centering
\includegraphics[width=0.95\textwidth,clip]{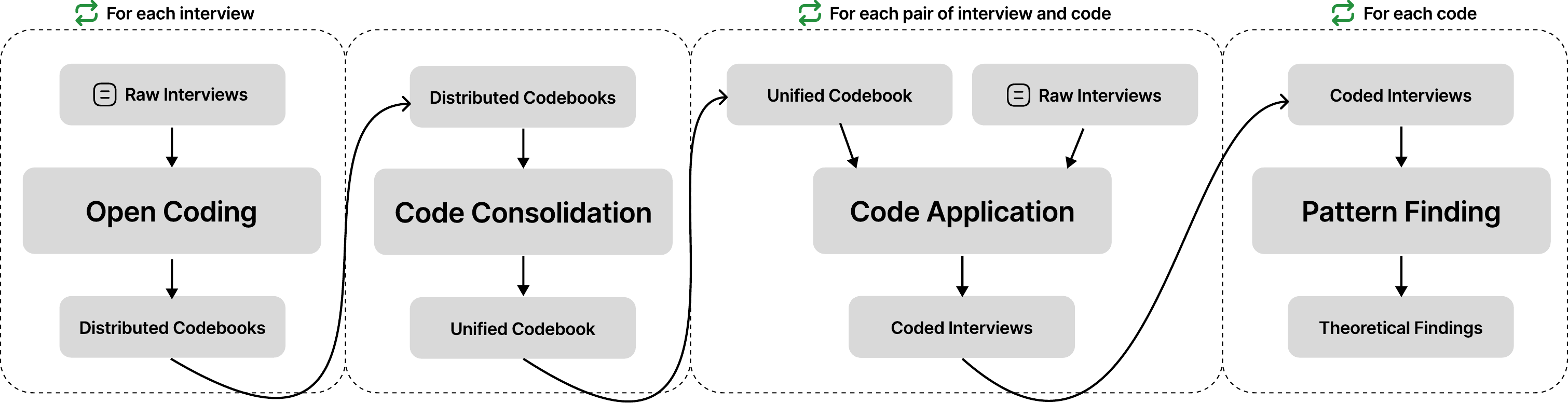}
\caption{Conceptual Illustration of the AI Co-Ethnographer Pipeline}
\label{fig:pipeline}
\end{figure*}

With the \textsc{AI Co-Ethnographer} (AICoE), we introduce a novel end-to-end pipeline that extends beyond the conventional focus on code assignments. The AICoE is part of a broader infrastructure for AI-assisted knowledge production, integrating diverse qualitative analysis methods, from open coding to pattern discovery. Whereas prior research has largely concentrated on automating the mapping of codes to text segments, our approach encompasses a more comprehensive qualitative analysis process. The pipeline extends the capabilities beyond the deductive application of pre-defined codes. Crucially, it also enables inductive code development and application, a process where novel codes are developed directly from the data itself instead of being pre-defined. 

\section{Related Research}
LLM development has spurred transdisciplinary efforts to automate scholarly work, especially qualitative textual analysis \cite{morgan_exploring_2023,petersen-frey_qualitative_2023,fischer_exploring_2024,lu_ai_2024, franken_ai_2025}, including ethnographically focused research \cite{lindgren_ai_2023}. This builds on a rich history of computational methods in qualitative research, from early tools like the General Inquirer \cite{stone_computer_1963} and Salton‘s vector space model \cite{salton_vector_1975}, to machine learning-based annotation \cite{sebastiani_machine_2002}, and open-source platforms like WordFreak \cite{morton_wordfreak_2003} and WebAnno \cite{yimam_automatic_2014}.
More recently, \citet{spinoso-di_piano_qualitative_2023} introduced the Qualitative Code Suggestion (QCS) task, which assists in coding by providing a ranked list of predefined codes for a given text passage. To evaluate QCS, the authors present CVDQuoding, an annotated dataset of interviews with women at risk of cardiovascular disease. Human evaluation shows that their system provides relevant suggestions, highlighting its potential as an assistive tool. However, limitations remain, including a focus on code assignment rather than full codebook development and a lack of evaluation in applied research settings.
Similarly, \citet{ziems_can_2024} evaluated the potential of LLMs for automating social science tasks, focusing on their zero-shot capabilities. Their findings indicate that LLMs demonstrate proficiency in both classification and explanation, suggesting their ability to augment the social science research pipeline. However, the authors do not recommend LLMs as a replacement for traditional methods.

\section{Methodology}

The \textsc{AI Co-Ethnographer} is composed of a comprehensive pipeline underpinned by LLMs to automate key qualitative research processes while aiming to preserve the interpretative depth central to ethnography. Building on recent advances in LLMs, the system mirrors several stages of qualitative analysis (see Figure \ref{fig:pipeline}): open coding, code consolidation, code application, and pattern finding. This approach enables scalable and consistent analysis of large volumes of qualitative data while mimicking ethnographic research practices.

\subsection{Open Coding}
 A first step can be called \textit{open coding}, where individual interviews are processed separately by the LLM. By isolating analyses per interview, the chosen research design addresses both the context window limitations of LLMs and the ethnographic principle of maintaining close connection to primary data. The system may suggest up to $N$ codes per interview, balancing descriptive and interpretive coding approaches and, in doing so, automating a time-consuming element of qualitative analysis.

\subsection{Code Consolidation}
The \textit{code consolidation} stage transitions to a global perspective and synthesizes findings across all interviews into a unified codebook. The synthesis process analyzes code overlap and merges similar concepts, culminating in a maximum of up to $M$ consolidated codes. This stage represents a crucial bridge between individual narratives and broader theoretical development, akin to manual axial coding but computationally scaled.
\subsection{Code Application}
The pipeline returns to a local perspective in the \textit{code application} stage, where each consolidated code is systematically applied to individual interview transcripts. Unlike existing approaches that work with limited text fragments \cite{spinoso-di_piano_qualitative_2023}, our system processes the entire interview for each code\footnote{We note that this approach allows for prompt caching for a more efficient application of the codes.}, thereby ensuring that the full conversational context informs the identification of relevant passages. This preserves the crucial ethnographic context of when and where statements occur. The system maintains connections between codes and input data through extracted text segments that can be mapped back to the original interviews, primarily via unique exact matches or substring matches. In rarer cases when such a match is unavailable, we instead rely on a sufficiently large word overlap measure using ROUGE \cite{lin_rouge_2004}, addressing both the technical need for systematic analysis and the ethnographic requirement for contextual grounding.
\subsection{Pattern Finding}
Finally, the \textit{pattern finding} stage shifts back to a holistic perspective, analyzing relationships between coded segments across the entire set of interviews to identify insights. This stage examines co-occurrence, contextual relationships, and thematic patterns, automating the transition from coding to broad theoretical and interpretative understanding.

\subsection{Prompt Engineering}
The developed prompts (see Appendix \ref{sec:appendix-prompts}) aim to emulate standard procedures in qualitative research, specifically in an ethnographic context. Each prompt corresponds to a phase of analysis and is structured to ensure methodological rigor. The scratchpad is important, as it allows the model to articulate its step-by-step reasoning, thereby making it transparent. By emphasizing verbatim text extraction and a strict correspondence between each extracted segment and the original interview line, we aim for high inter-rater reliability and transparency. Additionally, optional code descriptions during codebook development enhance clarity, and optional context helps guide the research direction. The \texttt{max\_codes} parameter is a technical restriction to avoid overly lengthy prompts, but in practice can be adjusted according to factors such as the model’s context length, its ability to maintain performance over long contexts, and the number of interviews. Although these prompts are illustrated with ethnographic interviews, the underlying principle of precise, code-based textual extraction readily extends to other qualitative research methodologies.
\section{Experiments and Results}

The system leverages \texttt{Llama-3.3-70B} \cite{dubey_llama_2024} as LLM, though the modular pipeline design permits integration with any modern LLM. We evaluate the model on three interviews each from the CVDQuoding and HiAICS datasets, the latter being our collection of interviews conducted as part of an ethnographic analysis with AI researchers. The study participants include both researchers who apply AI practically in their scientific disciplines and those who offer theoretical and critical analyses of AI's use in research. The interviews were transcribed using the speaker-attributed ASR system by \citet{nguyen_msa-asr_2025}.\footnote{We publish the HiAICS interviews under \url{https://codeberg.org/hiaics/interviews}.} 

\subsection{Semantic Relatedness of Codebooks}

To evaluate the semantic relatedness between different qualitative codebooks, we developed a novel framework for systematically comparing code taxonomies by specifying the following semantic relationships between codes:

\begin{itemize}
    \item \textbf{($M$) Match (1:1)} -- Defines codes capturing broadly similar concepts across codebooks, though they may use different terminology
    \item \textbf{($C$) Containment (1:n)} -- Indicates when one code represents a broader concept encompassing one or more codes from the other scheme
    \item \textbf{($P$) Partial Overlap (1:1)} -- Denotes codes that share some aspects of their meaning while maintaining distinct elements
    \item \textbf{($U$) Unmatched} -- Codes representing entirely unique aspects absent in the other codebook
\end{itemize}

A visual demonstration of these relations can be found in Figure \ref{fig:codebook-relations-viz} in the Appendix. Based on these relationships, we also developed a scoring method to quantify them. We normalize n:1 containments into atomic 1:1 relationships and assign weights for semantic relevance: $w_m = 1.0$ for matches, $w_c = 0.7$ for containments, and $w_p = 0.5$ for overlaps. For each code $x$, let $R(x)$ denote its set of relationships. The individual score $s(x)$, codebook scores $\tau_i$, and final score are calculated as:
\begin{align}
    s(x) &= \max(\{w_r : r \in R(x)\} \cup \{0\}) \\
    \tau_i &= \frac{1}{|i|} \sum_{x \in i} s(x) \quad \text{for } i \in \{A,B\} \\
    \tau_{sem} &= \frac{\tau_A + \tau_B}{2}
\end{align}

where $A$ and $B$ represent the two complete sets of codes in codebooks.

\begin{table}[h]
\centering
\scriptsize
\setlength{\tabcolsep}{0.105cm}
\begin{tabular}{llccccc}
\toprule
Schema 1 & Schema 2 & $M$ & $C$ & $P$ & $U$ & $\tau_{sem}$ \\
\midrule
Coder A & Coder B & 0.216 & 0.346 & 0.251 & 0.187 & \textbf{0.584} \\
Coder A & AICoE     & 0.206 & 0.480 & 0.191 & 0.123 & \textbf{0.638} \\
Coder B & AICoE     & 0.081 & 0.573 & 0.125 & 0.221 & \textbf{0.545} \\
\bottomrule
\end{tabular}
\caption{Distribution of relationship types comparing codebooks derived from the HiAICS dataset. A visual side-by-side comparison is provided in Figure~\ref{fig:codebook-comparison}, and detailed results in Table~\ref{tab:evaluator-scores-detailed} in the Appendix.}
\label{tab:schema-comparison}
\end{table}

\subsection{Relevance of Code Assignments}

To assess code-to-text relevance independently of upstream stages, we provided the system with \textit{human-curated codebooks} derived from prior manual analyses\footnote{Specifically, for the CVDQuoding dataset, which was published with two codebooks, we utilized Coder 2's codebook. For our HiAICS dataset, we employed a codebook developed by one of our expert annotators (Coder 1).}. This controlled setup isolates the code application mechanism. Several experts assessed whether human-assigned and AI-assigned codes were \textit{relevant} or \textit{irrelevant} to corresponding text segments, blinded to origin.

\begin{table}[h]
    \centering
    \scriptsize
    \begin{tabular}{lcc}
        \toprule
        Dataset & Human & AICoE \\
        \midrule
        CVDQuoding & 0.806 & 0.760 \\
        HiAICS & 0.740 & 0.560 \\
        \midrule
        Overall Average & \textbf{0.773} & \textbf{0.660} \\
        \bottomrule
    \end{tabular}
    \caption{Relevant code assignments averaged across interviews and evaluators, from human and AI coders; results for each evaluator are in Table \ref{tab:rel-code-assignments-detailed} in the Appendix}
    \label{tab:aggregated-metrics}
\end{table}

\subsection{Quality of Theoretical Findings}

To assess the quality of the generated findings, we conducted a human evaluation using three criteria:

\begin{itemize}
    \item ($G$) \textbf{Grounding} (\textit{Data Grounding, Evidence Support \& Accuracy}): Findings must be accurate, reliable, and well-supported by the interviews. Optimally, multiple coded segments are mentioned or provided.
    \item ($R$) \textbf{Relevance} (\textit{Alignment with Code \& Research Goals}): Findings should address the research objectives and the assigned code.
    \item ($I$) \textbf{Insight} (\textit{Insightfulness, Novelty \& Non-Triviality}): Findings should reveal deeper, non-obvious insights of intellectual value and avoid surface-level observations or trivialities.
\end{itemize}

For the HiAICS dataset, three experts who were asked to read the interviews before rated each finding on a 5-point Likert scale across these dimensions. The \%HQ metric (percentage of high-quality findings) reflects the proportion of codes yielding at least one finding with an average rating of 4.00 or higher across experts and criteria.

\begin{table}[h]
\centering
\scriptsize
\begin{tabular}{lccc}
\toprule
\textbf{} & \textbf{Mean} & \textbf{SD} & \textbf{\% HQ} \\
\midrule
Grounding & 3.42 & 0.61 & -- \\
Relevance & 3.76 & 0.41 & -- \\
Insight   & 3.29 & 0.46 & -- \\
\midrule
\textbf{Overall Quality} & \textbf{3.49} & \textbf{0.38} & \textbf{32.25} \\
\bottomrule
\end{tabular}
\caption{Evaluation scores for AICoE findings on HiAICS across 31 codes (151 total findings), detailed results for all findings are in Table \ref{tab:detailed-findings} and exemplary, high-quality findings are in Figure \ref{fig:high-quality-findings-examples}, both in the Appendix}
\label{tab:aggregated-pattern-evaluation}
\end{table}

\section{Discussion}

\paragraph{Alignments, Gaps, and New Perspectives in Codebooks.} The codebook alignments (Table~\ref{tab:schema-comparison}) indicate that AICoE is not meaningfully more divergent from either human-coded schema than the two human codebooks are from each other. However, a closer manual inspection of the codebooks reveals that AICoE tends to prioritize thematic concepts, whereas human coders occasionally add codes reflecting individual interviewee experiences (e.g., ``Biographical Context'' or ``Personal Work''). Notably, all three codebooks contained unique codes unmatched by the others, underscoring AICoE's potential to complement human analysis by offering alternative perspectives that can aid researchers in refining and expanding their codebooks.

\paragraph{Coding Performance Disparities.} The observed performance gap between human and AI coding in HiAICS ($\Delta = 0.180$) compared to CVDQuoding ($\Delta = 0.046$) presumably stems from inherent data characteristics. First and most importantly, CVDQuoding consists of structured interviews with predefined questions, likely providing clearer thematic boundaries that facilitate more consistent coding. Second, an interview in HiAICS contains, on average, approximately twice the word count (10,663 versus 5,163 words), increasing the complexity for the model to maintain contextual coherence. This aligns with previous evidence showing that LLM performance generally degrades as the context length increases \cite{liu_lost_2024}. Finally, the ASR-generated transcripts in HiAICS introduce linguistic noise through transcription artifacts and speech disfluencies.

\paragraph{Finding Meaning in Data.} The results in Table \ref{tab:aggregated-pattern-evaluation} underscore that AICoE reliably identifies theoretically relevant patterns, achieving an overall quality score of $3.49$ with $32.25\%$ of codes with high-quality findings ($\geq4.00$). Grounding ($3.42$) and relevance ($3.76$) outperformed insight ($3.29$), reflecting strength in anchoring findings in data and aligning them with research objectives while highlighting the difficulty of automating interpretative depth. Inter-rater correlations (see Appendix \ref{sec:corr_coff}) reveal more consistent assessments for grounding (E2–E3: $r = 0.6471$), but low agreement for relevance and insight (max $r = 0.1194$ and $0.2478$), indicating more subjective judgments in evaluating thematic alignment and the novelty of findings.

\paragraph{AI-Augmented Ethnography.} While our approach presents a systematic pipeline for qualitative analysis, it should not be viewed solely through the lens of automation. Rather, the framework embraces human expertise and allows for critical intervention at every stage. The \textit{unified codebook}, in particular, serves as a 'checkpoint' where researchers can review, refine, and adjust consolidated codes before proceeding to code application and pattern finding. Importantly, our framework also supports \textit{deductive coding} approaches, allowing researchers to bypass the open coding and code consolidation stages by directly applying a pre-existing or theory-driven codebook. This flexibility extends throughout the pipeline – researchers can iterate through stages multiple times, run parallel samples, or modify intermediate outputs as needed. The pattern finding stage, as a final step, exemplifies this collaboration, where computational analysis assists human insight rather than replaces it. 

\paragraph{Tool, Partner, or Epistemic Medium?} Based on these considerations, it is imperative to clarify that the AI Co-Ethnographer is conceptualized neither as a mere instrument nor as a quasi-human agent. We must conscientiously avoid both anthropomorphic and anthropocentric framings, and equally guard against its reduction to a static, pre-determined technological artifact. Rather, we posit the AI Co-Ethnographer as an epistemic medium, one that facilitates and supports the generation of knowledge, while remaining subject to critical reflection.
Serving as such a medium, the AI Co-Ethnographer enriches the research infrastructure that underpins ethnographic and, more comprehensively, qualitative research.

\paragraph{Multimodality and Data Heterogeneity.} Future research must address the inherent multimodality and data heterogeneity of scientific processes related to the analysis of qualitative data. While our pipeline focuses on textual data (interview transcripts), scientific activity extends far beyond text. It encompasses diverse multimodal inputs or media: spoken language (interviews, lectures, meetings), visual elements (slides, graphics, videos), and discipline-specific sensor data \cite{yang_visual_1998,bett_multimodal_2000}. Scientific discussions, for instance, exemplify this multimodality, integrating spoken interaction, nonverbal cues like gesture and gaze, or the presentation of visual materials. Achieving a broader, faster, and more contextualized understanding of scientific processes requires developing methods to process, interpret, and synthesize these diverse, cross-modal signals.

\section{Conclusion}

The AI Co-Ethnographer demonstrates both the potential and limitations of AI-supported qualitative research. Our evaluation reveals robust codebook development, reasonable code assignments, and the ability to generate meaningful findings. This represents a promising direction for qualitative research, enabling the processing of large volumes of data while maintaining analytical depth. Beyond functioning as a mere tool, AICoE serves as an epistemic medium in the research process.

\section*{Limitations}
Debates continue over the extent to which ethnographic approaches to qualitative research can be automated or delegated to AI systems. However, larger amounts of ethnographic data can only be analyzed with the support of corresponding systems. In the context of our research, every phase of qualitative data analysis remains intrinsically tied to ethnographic experience and observation of human subjects.
Future refinements to our framework could prioritize the specificities inherent in ethnographic data analysis, placing them at the core of this epistemic conduit. For instance, we might contemplate a more nuanced synthesis of interview transcripts and observational records, such as field notes. However, we consider it an asset, rather than a liability, that this proposed epistemic conduit offers flexible support for the annotation and interpretation of qualitative research data beyond solely ethnographic contexts. Consequently, it has the potential to reshape how AI supports transdisciplinary qualitative research in the future.

\section*{Ethics}
The use of LLMs for automatic coding and qualitative analysis of research materials involves ethical challenges related to data privacy, algorithmic biases, and transparency. Researchers should ensure that participant data is adequately protected and obtain their informed consent for AI-assisted analysis. It is essential to critically evaluate potential biases in LLM-generated annotations and interpretations and to ensure transparency in AI’s role in the analytical process. Clear authorship and accountability guidelines are necessary for LLM-assisted qualitative analysis. Finally, it is important to balance leveraging AI’s ability to handle massive datasets with maintaining rigorous ethical research standards.

\section*{Acknowledgments}
This research is supported by the project ``How is AI Changing Science? Research in the Era of Learning Algorithms'' (HiAICS), funded by the Volkswagen Foundation. We thank Matthias Ernst, Christine Hämmerling, Birte Luisa Kuhle, Markus Ramsauer, Johanna Maria Toussaint, and  Charmaine Voigt for their contributions to annotation and evaluation.

\bibliography{custom}

\begin{thebibliography}{40}
\providecommand{\natexlab}[1]{#1}

\bibitem[{Bernard et~al.(2016)Bernard, Wutich, and Ryan}]{bernard_analyzing_2016}
H.~Russell Bernard, Amber Wutich,  and Gery~W. Ryan. 2016.
\newblock \emph{Analyzing {Qualitative} {Data}: {Systematic} {Approaches}}.
\newblock SAGE Publications.
\newblock Google-Books-ID: yAi1DAAAQBAJ.

\bibitem[{Bernard(2011)}]{bernard_research_2011}
Harvey~Russell Bernard. 2011.
\newblock \emph{Research {Methods} in {Anthropology}: {Qualitative} and {Quantitative} {Approaches}}.
\newblock Rowman Altamira.

\bibitem[{Bett et~al.(2000)Bett, Gross, Yu, Zhu, Pan, Yang, and Waibel}]{bett_multimodal_2000}
Michael Bett, Ralph Gross, Hua Yu, Xiaojin Zhu, Yue Pan, Jie Yang,  and Alex Waibel. 2000.
\newblock Multimodal {Meeting} {Tracker}.
\newblock In \emph{{RIAO}}, pages 32--45. Paris, France.

\bibitem[{Creswell(2015)}]{creswell_30_2015}
John~W. Creswell. 2015.
\newblock \emph{30 {Essential} {Skills} for the {Qualitative} {Researcher}}.
\newblock SAGE Publications.
\newblock Google-Books-ID: fkJsCgAAQBAJ.

\bibitem[{Creswell and Poth(2017)}]{creswell_qualitative_2017}
John~W. Creswell  and Cheryl~N. Poth. 2017.
\newblock \emph{Qualitative {Inquiry} and {Research} {Design}: {Choosing} {Among} {Five} {Approaches}}, 4 edition.
\newblock SAGE Publications, Inc.

\bibitem[{Denzin et~al.(2023)Denzin, Lincoln, Giardina, and Cannella}]{denzin_sage_2023}
Norman~K. Denzin, Yvonna~S. Lincoln, Michael~Donald Giardina,  and Gaile~S. Cannella. 2023.
\newblock \emph{The {SAGE} {Handbook} of {Qualitative} {Research}}, 6 edition.
\newblock SAGE Publications, Inc, Los Angeles London New Delhi Singapore Washington DC Melbourne.

\bibitem[{Dippel and Sudmann(2023)}]{lindgren_ai_2023}
Anne Dippel  and Andreas Sudmann. 2023.
\newblock \href {https://doi.org/10.4337/9781803928562.00083} {{AI} ethnography}.
\newblock In Simon Lindgren, editor, \emph{Handbook of {Critical} {Studies} of {Artificial} {Intelligence}}, pages 826--844. Edward Elgar Publishing.

\bibitem[{Dubey et~al.(2024)Dubey, Jauhri, Pandey, Kadian, Al-Dahle, Letman, Mathur, Schelten, Vaughan, Yang, Fan, Goyal, Hartshorn, Yang, Mitra, Sravankumar, Korenev, Hinsvark, Rao, Zhang, Rodriguez, Gregerson, Spataru, Roziere, Biron, Tang, Chern, Caucheteux, Nayak, Bi, Marra, McConnell, Keller, Touret, Wu, Wong, Ferrer, Nikolaidis, Allonsius, Song, Pintz, Livshits, Wyatt, Esiobu, Choudhary, Mahajan, Garcia-Olano, Perino, Hupkes, Lakomkin, AlBadawy, Lobanova, Dinan, Smith, Radenovic, Guzmán, Zhang, Synnaeve, Lee, Anderson, Thattai, Nail, Mialon, Pang, Cucurell, Nguyen, Korevaar, Xu, Touvron, Zarov, Ibarra, Kloumann, Misra, Evtimov, Zhang, Copet, Lee, Geffert, Vranes, Park, Mahadeokar, Shah, Linde, Billock, Hong, Lee, Fu, Chi, Huang, Liu, Wang, Yu, Bitton, Spisak, Park, Rocca, Johnstun, Saxe, Jia, Alwala, Prasad, Upasani, Plawiak, Li, Heafield, Stone, El-Arini, Iyer, Malik, Chiu, Bhalla, Lakhotia, Rantala-Yeary, Maaten, Chen, Tan, Jenkins, Martin, Madaan, Malo, Blecher, Landzaat, Oliveira, Muzzi,
  Pasupuleti, Singh, Paluri, Kardas, Tsimpoukelli, Oldham, Rita, Pavlova, Kambadur, Lewis, Si, Singh, Hassan, Goyal, Torabi, Bashlykov, Bogoychev, Chatterji, Zhang, Duchenne, Çelebi, Alrassy, Zhang, Li, Vasic, Weng, Bhargava, Dubal, Krishnan, Koura, Xu, He, Dong, Srinivasan, Ganapathy, Calderer, Cabral, Stojnic, Raileanu, Maheswari, Girdhar, Patel, Sauvestre, Polidoro, Sumbaly, Taylor, Silva, Hou, Wang, Hosseini, Chennabasappa, Singh, Bell, Kim, Edunov, Nie, Narang, Raparthy, Shen, Wan, Bhosale, Zhang, Vandenhende, Batra, Whitman, Sootla, Collot, Gururangan, Borodinsky, Herman, Fowler, Sheasha, Georgiou, Scialom, Speckbacher, Mihaylov, Xiao, Karn, Goswami, Gupta, Ramanathan, Kerkez, Gonguet, Do, Vogeti, Albiero, Petrovic, Chu, Xiong, Fu, Meers, Martinet, Wang, Wang, Tan, Xia, Xie, Jia, Wang, Goldschlag, Gaur, Babaei, Wen, Song, Zhang, Li, Mao, Coudert, Yan, Chen, Papakipos, Singh, Srivastava, Jain, Kelsey, Shajnfeld, Gangidi, Victoria, Goldstand, Menon, Sharma, Boesenberg, Baevski, Feinstein, Kallet,
  Sangani, Teo, Yunus, Lupu, Alvarado, Caples, Gu, Ho, Poulton, Ryan, Ramchandani, Dong, Franco, Goyal, Saraf, Chowdhury, Gabriel, Bharambe, Eisenman, Yazdan, James, Maurer, Leonhardi, Huang, Loyd, Paola, Paranjape, Liu, Wu, Ni, Hancock, Wasti, Spence, Stojkovic, Gamido, Montalvo, Parker, Burton, Mejia, Liu, Wang, Kim, Zhou, Hu, Chu, Cai, Tindal, Feichtenhofer, Gao, Civin, Beaty, Kreymer, Li, Adkins, Xu, Testuggine, David, Parikh, Liskovich, Foss, Wang, Le, Holland, Dowling, Jamil, Montgomery, Presani, Hahn, Wood, Le, Brinkman, Arcaute, Dunbar, Smothers, Sun, Kreuk, Tian, Kokkinos, Ozgenel, Caggioni, Kanayet, Seide, Florez, Schwarz, Badeer, Swee, Halpern, Herman, Sizov, Guangyi, Zhang, Lakshminarayanan, Inan, Shojanazeri, Zou, Wang, Zha, Habeeb, Rudolph, Suk, Aspegren, Goldman, Zhan, Damlaj, Molybog, Tufanov, Leontiadis, Veliche, Gat, Weissman, Geboski, Kohli, Lam, Asher, Gaya, Marcus, Tang, Chan, Zhen, Reizenstein, Teboul, Zhong, Jin, Yang, Cummings, Carvill, Shepard, McPhie, Torres, Ginsburg, Wang, Wu, U,
  Saxena, Khandelwal, Zand, Matosich, Veeraraghavan, Michelena, Li, Jagadeesh, Huang, Chawla, Huang, Chen, Garg, A, Silva, Bell, Zhang, Guo, Yu, Moshkovich, Wehrstedt, Khabsa, Avalani, Bhatt, Mankus, Hasson, Lennie, Reso, Groshev, Naumov, Lathi, Keneally, Liu, Seltzer, Valko, Restrepo, Patel, Vyatskov, Samvelyan, Clark, Macey, Wang, Hermoso, Metanat, Rastegari, Bansal, Santhanam, Parks, White, Bawa, Singhal, Egebo, Usunier, Mehta, Laptev, Dong, Cheng, Chernoguz, Hart, Salpekar, Kalinli, Kent, Parekh, Saab, Balaji, Rittner, Bontrager, Roux, Dollar, Zvyagina, Ratanchandani, Yuvraj, Liang, Alao, Rodriguez, Ayub, Murthy, Nayani, Mitra, Parthasarathy, Li, Hogan, Battey, Wang, Howes, Rinott, Mehta, Siby, Bondu, Datta, Chugh, Hunt, Dhillon, Sidorov, Pan, Mahajan, Verma, Yamamoto, Ramaswamy, Lindsay, Lindsay, Feng, Lin, Zha, Patil, Shankar, Zhang, Zhang, Wang, Agarwal, Sajuyigbe, Chintala, Max, Chen, Kehoe, Satterfield, Govindaprasad, Gupta, Deng, Cho, Virk, Subramanian, Choudhury, Goldman, Remez, Glaser, Best,
  Koehler, Robinson, Li, Zhang, Matthews, Chou, Shaked, Vontimitta, Ajayi, Montanez, Mohan, Kumar, Mangla, Ionescu, Poenaru, Mihailescu, Ivanov, Li, Wang, Jiang, Bouaziz, Constable, Tang, Wu, Wang, Wu, Gao, Kleinman, Chen, Hu, Jia, Qi, Li, Zhang, Zhang, Adi, Nam, Yu, Wang, Zhao, Hao, Qian, Li, He, Rait, DeVito, Rosnbrick, Wen, Yang, Zhao, and Ma}]{dubey_llama_2024}
Abhimanyu Dubey, Abhinav Jauhri, Abhinav Pandey, Abhishek Kadian, Ahmad Al-Dahle, Aiesha Letman, Akhil Mathur, Alan Schelten, Alex Vaughan, Amy Yang, Angela Fan, Anirudh Goyal, Anthony Hartshorn, Aobo Yang, Archi Mitra, Archie Sravankumar, Artem Korenev, Arthur Hinsvark, Arun Rao, ...,  and Zhiyu Ma. 2024.
\newblock \href {https://doi.org/10.48550/arXiv.2407.21783} {The {Llama} 3 {Herd} of {Models}}.
\newblock \emph{arXiv preprint}.
\newblock ArXiv:2407.21783 [cs].

\bibitem[{Elliott(2018)}]{elliott_thinking_2018}
V.~Elliott. 2018.
\newblock \href {https://ora.ox.ac.uk/objects/uuid:5304bf7f-6214-4939-9f1b-b64415d4fac1} {Thinking about the coding process in qualitative data analysis}.
\newblock \emph{Qualitative Report}, 23(11).
\newblock Publisher: Nova Southeastern University.

\bibitem[{Fischer and Biemann(2024)}]{fischer_exploring_2024}
Tim Fischer  and Chris Biemann. 2024.
\newblock \href {https://doi.org/10.18653/v1/2024.nlp4dh-1.41} {Exploring {Large} {Language} {Models} for {Qualitative} {Data} {Analysis}}.
\newblock In \emph{Proceedings of the 4th {International} {Conference} on {Natural} {Language} {Processing} for {Digital} {Humanities}}, pages 423--437, Miami, USA. Association for Computational Linguistics.

\bibitem[{Franken and Vepřek(2025)}]{franken_ai_2025}
Lina Franken  and Libuše~Hannah Vepřek. 2025.
\newblock \href {https://doi.org/10.1007/978-3-658-08460-8_98-1} {{AI} in and for {Qualitative} {Research}}.
\newblock In Heidrun Friese, Marcus Nolden,  and Miriam Schreiter, editors, \emph{Handbuch {Soziale} {Praktiken} und {Digitale} {Alltagswelten}}, pages 1--9. Springer Fachmedien, Wiesbaden.

\bibitem[{Harding(2013)}]{harding_qualitative_2013}
Jamie Harding. 2013.
\newblock \emph{Qualitative {Data} {Analysis} from {Start} to {Finish}}.
\newblock SAGE.
\newblock Google-Books-ID: 9YUQAgAAQBAJ.

\bibitem[{Holton(2007)}]{holton_coding_2007}
Judith~A. Holton. 2007.
\newblock \href {https://www.torrossa.com/gs/resourceProxy?an=5019441&publisher=FZ7200#page=298} {The coding process and its challenges}.
\newblock \emph{The Sage handbook of grounded theory}, 3:265--289.

\bibitem[{Hori et~al.(2002)Hori, Furui, Malkin, Yu, and Waibel}]{hori_automatic_2002}
Chiori Hori, Sadaoki Furui, Rob Malkin, Hua Yu,  and Alex Waibel. 2002.
\newblock Automatic speech summarization applied to {English} broadcast news speech.
\newblock In \emph{2002 {IEEE} {International} {Conference} on {Acoustics}, {Speech}, and {Signal} {Processing}}, volume~1, pages I--9. IEEE.

\bibitem[{Lin(2004)}]{lin_rouge_2004}
Chin-Yew Lin. 2004.
\newblock \href {https://aclanthology.org/W04-1013/} {{ROUGE}: {A} {Package} for {Automatic} {Evaluation} of {Summaries}}.
\newblock In \emph{Text {Summarization} {Branches} {Out}}, pages 74--81, Barcelona, Spain. Association for Computational Linguistics.

\bibitem[{Liu et~al.(2024)Liu, Lin, Hewitt, Paranjape, Bevilacqua, Petroni, and Liang}]{liu_lost_2024}
Nelson~F. Liu, Kevin Lin, John Hewitt, Ashwin Paranjape, Michele Bevilacqua, Fabio Petroni,  and Percy Liang. 2024.
\newblock \href {https://doi.org/10.1162/tacl_a_00638} {Lost in the {Middle}: {How} {Language} {Models} {Use} {Long} {Contexts}}.
\newblock \emph{Transactions of the Association for Computational Linguistics}, 12:157--173.
\newblock Place: Cambridge, MA Publisher: MIT Press.

\bibitem[{Lu et~al.(2024)Lu, Lu, Lange, Foerster, Clune, and Ha}]{lu_ai_2024}
Chris Lu, Cong Lu, Robert~Tjarko Lange, Jakob Foerster, Jeff Clune,  and David Ha. 2024.
\newblock \href {https://doi.org/10.48550/arXiv.2408.06292} {The {AI} {Scientist}: {Towards} {Fully} {Automated} {Open}-{Ended} {Scientific} {Discovery}}.
\newblock \emph{arXiv preprint}.
\newblock ArXiv:2408.06292 [cs].

\bibitem[{Miles et~al.(2019)Miles, Huberman, and Saldana}]{miles_qualitative_2019}
Matthew~B. Miles, A.~Michael Huberman,  and Johnny Saldana. 2019.
\newblock \emph{Qualitative {Data} {Analysis}: {A} {Methods} {Sourcebook}}.
\newblock SAGE Publications.
\newblock Google-Books-ID: Bt0uuQEACAAJ.

\bibitem[{Morgan(2023)}]{morgan_exploring_2023}
David~L. Morgan. 2023.
\newblock \href {https://doi.org/10.1177/16094069231211248} {Exploring the {Use} of {Artificial} {Intelligence} for {Qualitative} {Data} {Analysis}: {The} {Case} of {ChatGPT}}.
\newblock \emph{International Journal of Qualitative Methods}, 22:16094069231211248.
\newblock Publisher: SAGE Publications Inc.

\bibitem[{Morton and LaCivita(2003)}]{morton_wordfreak_2003}
Thomas Morton  and Jeremy LaCivita. 2003.
\newblock \href {https://aclanthology.org/N03-4009/} {{WordFreak}: {An} {Open} {Tool} for {Linguistic} {Annotation}}.
\newblock In \emph{Companion {Volume} of the {Proceedings} of {HLT}-{NAACL} 2003 - {Demonstrations}}, pages 17--18.

\bibitem[{Nguyen and Waibel(2025)}]{nguyen_msa-asr_2025}
Thai-Binh Nguyen  and Alexander Waibel. 2025.
\newblock \href {https://doi.org/10.48550/arXiv.2411.18152} {{MSA}-{ASR}: {Efficient} {Multilingual} {Speaker} {Attribution} with frozen {ASR} {Models}}.
\newblock \emph{arXiv preprint}.
\newblock ArXiv:2411.18152 [cs].

\bibitem[{Nguyen et~al.(2021)Nguyen, Stüker, and Waibel}]{nguyen_super-human_2021}
Thai-Son Nguyen, Sebastian Stüker,  and Alex Waibel. 2021.
\newblock \href {https://doi.org/10.21437/Interspeech.2021-1114} {Super-{Human} {Performance} in {Online} {Low}-{Latency} {Recognition} of {Conversational} {Speech}}.
\newblock In \emph{Interspeech 2021}, pages 1762--1766.
\newblock ISSN: 2958-1796.

\bibitem[{Petersen-Frey et~al.(2023)Petersen-Frey, Fischer, Schneider, Eiser, Koch, and Biemann}]{petersen-frey_qualitative_2023}
Fynn Petersen-Frey, Tim Fischer, Florian Schneider, Isabel Eiser, Gertraud Koch,  and Chris Biemann. 2023.
\newblock \href {https://aclanthology.org/2023.konvens-main.5/} {From {Qualitative} to {Quantitative} {Research}: {Semi}-{Automatic} {Annotation} {Scaling} in the {Digital} {Humanities}}.
\newblock In \emph{Proceedings of the 19th {Conference} on {Natural} {Language} {Processing} ({KONVENS} 2023)}, pages 52--62, Ingolstadt, Germany. Association for Computational Lingustics.

\bibitem[{Rasheed et~al.(2024)Rasheed, Waseem, Ahmad, Kemell, Xiaofeng, Duc, and Abrahamsson}]{rasheed_can_2024}
Zeeshan Rasheed, Muhammad Waseem, Aakash Ahmad, Kai-Kristian Kemell, Wang Xiaofeng, Anh~Nguyen Duc,  and Pekka Abrahamsson. 2024.
\newblock \href {https://doi.org/10.48550/arXiv.2402.01386} {Can {Large} {Language} {Models} {Serve} as {Data} {Analysts}? {A} {Multi}-{Agent} {Assisted} {Approach} for {Qualitative} {Data} {Analysis}}.
\newblock \emph{arXiv preprint}.
\newblock ArXiv:2402.01386 [cs].

\bibitem[{Retkowski and Waibel(2024{\natexlab{a}})}]{retkowski_text_2024}
Fabian Retkowski  and Alexander Waibel. 2024{\natexlab{a}}.
\newblock \href {https://aclanthology.org/2024.eacl-long.25/} {From {Text} {Segmentation} to {Smart} {Chaptering}: {A} {Novel} {Benchmark} for {Structuring} {Video} {Transcriptions}}.
\newblock In \emph{Proceedings of the 18th {Conference} of the {European} {Chapter} of the {Association} for {Computational} {Linguistics} ({Volume} 1: {Long} {Papers})}, pages 406--419, St. Julian's, Malta. Association for Computational Linguistics.

\bibitem[{Retkowski and Waibel(2024{\natexlab{b}})}]{retkowski_zero-shot_2024}
Fabian Retkowski  and Alexander Waibel. 2024{\natexlab{b}}.
\newblock \href {https://doi.org/10.48550/arXiv.2501.00233} {Zero-{Shot} {Strategies} for {Length}-{Controllable} {Summarization}}.
\newblock \emph{arXiv preprint}.
\newblock ArXiv:2501.00233 [cs].

\bibitem[{Ritchie et~al.(2014)Ritchie, Lewis, Nicholls, and Ormston}]{ritchie_qualitative_2014}
Jane Ritchie, Jane Lewis, Carol~McNaughton Nicholls,  and Rachel Ormston. 2014.
\newblock \emph{Qualitative {Research} {Practice}: {A} {Guide} for {Social} {Science} {Students} and {Researchers}}.
\newblock SAGE Publications.
\newblock Google-Books-ID: zkITlwEACAAJ.

\bibitem[{Saldana(2015)}]{saldana_coding_2015}
Johnny Saldana. 2015.
\newblock \emph{The {Coding} {Manual} for {Qualitative} {Researchers}}.
\newblock SAGE.
\newblock Google-Books-ID: ZhxiCgAAQBAJ.

\bibitem[{Salton et~al.(1975)Salton, Wong, and Yang}]{salton_vector_1975}
G.~Salton, A.~Wong,  and C.~S. Yang. 1975.
\newblock \href {https://doi.org/10.1145/361219.361220} {A vector space model for automatic indexing}.
\newblock \emph{Communications of the ACM}, 18(11):613--620.

\bibitem[{Sebastiani(2002)}]{sebastiani_machine_2002}
Fabrizio Sebastiani. 2002.
\newblock \href {https://doi.org/10.1145/505282.505283} {Machine learning in automated text categorization}.
\newblock \emph{ACM Computing Surveys}, 34(1):1--47.

\bibitem[{Singhal et~al.(2025)Singhal, Tu, Gottweis, Sayres, Wulczyn, Amin, Hou, Clark, Pfohl, Cole-Lewis, Neal, Rashid, Schaekermann, Wang, Dash, Chen, Shah, Lachgar, Mansfield, Prakash, Green, Dominowska, Agüera~y Arcas, Tomašev, Liu, Wong, Semturs, Mahdavi, Barral, Webster, Corrado, Matias, Azizi, Karthikesalingam, and Natarajan}]{singhal_toward_2025}
Karan Singhal, Tao Tu, Juraj Gottweis, Rory Sayres, Ellery Wulczyn, Mohamed Amin, Le~Hou, Kevin Clark, Stephen~R. Pfohl, Heather Cole-Lewis, Darlene Neal, Qazi~Mamunur Rashid, Mike Schaekermann, Amy Wang, Dev Dash, Jonathan~H. Chen, Nigam~H. Shah, Sami Lachgar, Philip~Andrew Mansfield, ...,  and Vivek Natarajan. 2025.
\newblock \href {https://doi.org/10.1038/s41591-024-03423-7} {Toward expert-level medical question answering with large language models}.
\newblock \emph{Nature Medicine}, pages 1--8.
\newblock Publisher: Nature Publishing Group.

\bibitem[{Spinoso-Di~Piano et~al.(2023)Spinoso-Di~Piano, Rahimi, and Cheung}]{spinoso-di_piano_qualitative_2023}
Cesare Spinoso-Di~Piano, Samira Rahimi,  and Jackie Cheung. 2023.
\newblock \href {https://doi.org/10.18653/v1/2023.findings-emnlp.993} {Qualitative {Code} {Suggestion}: {A} {Human}-{Centric} {Approach} to {Qualitative} {Coding}}.
\newblock In \emph{Findings of the {Association} for {Computational} {Linguistics}: {EMNLP} 2023}, pages 14887--14909, Singapore. Association for Computational Linguistics.

\bibitem[{Stone and Hunt(1963)}]{stone_computer_1963}
Philip~J. Stone  and Earl~B. Hunt. 1963.
\newblock \href {https://doi.org/10.1145/1461551.1461583} {A computer approach to content analysis: studies using the {General} {Inquirer} system}.
\newblock In \emph{Proceedings of the {May} 21-23, 1963, spring joint computer conference on - {AFIPS} '63 ({Spring})}, page 241, Detroit, Michigan. ACM Press.

\bibitem[{Tuschling et~al.(2023)Tuschling, Sudmann, and Dotzler}]{tuschling_chatgpt_2023}
Anna Tuschling, Andreas Sudmann,  and Bernhard~J. Dotzler, editors. 2023.
\newblock \href {https://doi.org/10.14361/9783839469088} {\emph{{ChatGPT} und andere »{Quatschmaschinen}«: {Gespräche} mit {Künstlicher} {Intelligenz}}}.
\newblock transcript Verlag.
\newblock Accepted: 2024-02-02T16:04:26Z.

\bibitem[{Xi et~al.(2023)Xi, Chen, Guo, He, Ding, Hong, Zhang, Wang, Jin, Zhou, Zheng, Fan, Wang, Xiong, Zhou, Wang, Jiang, Zou, Liu, Yin, Dou, Weng, Cheng, Zhang, Qin, Zheng, Qiu, Huang, and Gui}]{xi_rise_2023}
Zhiheng Xi, Wenxiang Chen, Xin Guo, Wei He, Yiwen Ding, Boyang Hong, Ming Zhang, Junzhe Wang, Senjie Jin, Enyu Zhou, Rui Zheng, Xiaoran Fan, Xiao Wang, Limao Xiong, Yuhao Zhou, Weiran Wang, Changhao Jiang, Yicheng Zou, Xiangyang Liu, ...,  and Tao Gui. 2023.
\newblock \href {https://doi.org/10.48550/arXiv.2309.07864} {The {Rise} and {Potential} of {Large} {Language} {Model} {Based} {Agents}: {A} {Survey}}.
\newblock \emph{arXiv preprint}.
\newblock ArXiv:2309.07864 [cs].

\bibitem[{Yang et~al.(1998)Yang, Stiefelhagen, Meier, and Waibel}]{yang_visual_1998}
Jie Yang, Rainer Stiefelhagen, Uwe Meier,  and Alex Waibel. 1998.
\newblock Visual tracking for multimodal human computer interaction.
\newblock In \emph{Proceedings of the {SIGCHI} conference on {Human} factors in computing systems}, pages 140--147.

\bibitem[{Yimam et~al.(2014)Yimam, Biemann, De~Castilho, and Gurevych}]{yimam_automatic_2014}
Seid~Muhie Yimam, Chris Biemann, Richard~Eckart De~Castilho,  and Iryna Gurevych. 2014.
\newblock \href {https://aclanthology.org/P14-5016.pdf} {Automatic annotation suggestions and custom annotation layers in {WebAnno}}.
\newblock In \emph{Proceedings of 52nd annual meeting of the {Association} for {Computational} {Linguistics}: {System} {Demonstrations}}, pages 91--96.

\bibitem[{Zechner and Waibel(2000)}]{zechner_diasumm_2000}
Klaus Zechner  and Alex Waibel. 2000.
\newblock \href {https://aclanthology.org/C00-2140/} {{DIASUMM}: {Flexible} {Summarization} of {Spontaneous} {Dialogues} in {Unrestricted} {Domains}}.
\newblock In \emph{{COLING} 2000 {Volume} 2: {The} 18th {International} {Conference} on {Computational} {Linguistics}}.

\bibitem[{Zhang et~al.(2024)Zhang, Ladhak, Durmus, Liang, McKeown, and Hashimoto}]{zhang_benchmarking_2024}
Tianyi Zhang, Faisal Ladhak, Esin Durmus, Percy Liang, Kathleen McKeown,  and Tatsunori~B. Hashimoto. 2024.
\newblock \href {https://doi.org/10.1162/tacl_a_00632} {Benchmarking {Large} {Language} {Models} for {News} {Summarization}}.
\newblock \emph{Transactions of the Association for Computational Linguistics}, 12:39--57.
\newblock Place: Cambridge, MA Publisher: MIT Press.

\bibitem[{Ziems et~al.(2024)Ziems, Held, Shaikh, Chen, Zhang, and Yang}]{ziems_can_2024}
Caleb Ziems, William Held, Omar Shaikh, Jiaao Chen, Zhehao Zhang,  and Diyi Yang. 2024.
\newblock \href {https://doi.org/10.1162/coli_a_00502} {Can {Large} {Language} {Models} {Transform} {Computational} {Social} {Science}?}
\newblock \emph{Computational Linguistics}, 50(1):237--291.

\end{thebibliography}

\newpage

\FloatBarrier

\newpage

\clearpage

\appendix

\section{Pipeline Prompts}
\label{sec:appendix-prompts}

\noindent\begin{minipage}{\textwidth}
\begin{promptbox}[title={Open Coding}]

\small

You are an AI assistant tasked with suggesting relevant codes for an ethnographic interview transcript. In ethnography, coding is the process of assigning labels or categories to segments of qualitative data to identify themes and patterns. This is a crucial step in analyzing interview data.\\

You will be presented with a transcript from an ethnographic interview. Your task is to suggest a set of codes that are relevant to this transcript. Remember, you are not assigning codes to specific sentences but rather proposing a list of codes that could be used to analyze this transcript.\\

Here is the transcript:\\

<transcript>\\
\param{transcript}\\
</transcript>\\

Please analyze this transcript and suggest a set of codes that could be used to categorize and understand the themes present in the interview. Follow these guidelines:
\begin{enumerate}[nosep,itemsep=2pt,topsep=4pt]
    \item Codes should be concise, typically consisting of one to three words
    \item Codes should capture key concepts, themes, or ideas present in the transcript
    \item Aim for a mix of descriptive codes (what is happening) and interpretive codes (the underlying meaning)
    \item Consider both explicit content and implicit meanings in the transcript
    \item Avoid overly broad or vague codes
    \item You are free to suggest up to \param{max\_codes} codes, depending on the complexity and length of the transcript
    \item[\textcolor{optional}{7.}] \optionalinline{Provide a brief description (up to 20 words) for each code to clarify its meaning and application that differentiates it from other codes.}
\end{enumerate}
\vspace{6pt}
\optional{You will be provided context that you can and should consider when suggesting codes.\\[1em]
<context>\\
\param{context}\\
</context>}\leavevmode\\
\\[1em]
Before providing your final list of codes, use the <scratchpad> to think through your process:\\

<scratchpad>
\begin{enumerate}[nosep,itemsep=2pt,topsep=4pt]
\item Identify the main topics discussed in the interview
\item Note any recurring themes or ideas
\item Consider the context and any underlying meanings
\item Think about the interviewee's experiences, attitudes, and behaviors
\item Reflect on how these elements could be categorized into codes
\end{enumerate}
</scratchpad>\\

Now, please provide your suggested list of codes for this transcript. Present your codes in the following format:\\

<suggested\_codes>
\begin{itemize}[nosep,itemsep=2pt,topsep=4pt]
    \item Code 1 \optionalinline{| Description that explains the meaning and context of Code 1 in up to 20 words}
    \item Code 2 \optionalinline{| Description that explains the meaning and context of Code 2 in up to 20 words}
    \item[] ...
\end{itemize}
</suggested\_codes>\\

Remember, these codes should be relevant to the given transcript and useful for further analysis in an ethnographic study. Do not write content outside <scratchpad> or <suggested\_codes>.

\promptsep

\paramheading

\begin{paramlist}
    \item \param{transcript}: The raw interview transcript to analyze
    \item \param{context}: Optional additional context to consider
    \item \param{max\_codes}: Maximum number of codes to suggest
\end{paramlist}
\end{promptbox}
\end{minipage}

\newpage

\newpage

\begin{figure*}
\begin{promptbox}[title={Code Consolidation}]

\small

You are an AI assistant tasked with generating a comprehensive set of codes based on multiple ethnographic interviews. Your goal is to create a coherent and inclusive set of codes that covers the themes from all the interviews provided.\\

\optional{You will be provided context that you can and should consider when creating your final set of codes.\\[1em]
<context>\\
\param{context}\\
</context>}\leavevmode\\[1em]

You will be presented with sets of codes generated from multiple interviews. These codes are contained in the following variable:\\

\noindent
<interview\_codes>\\
\param{interview\_codes}\\
</interview\_codes>\\

Analyze these sets of codes and create a single, comprehensive set that encompasses the themes from all interviews. Follow these guidelines:
\begin{enumerate}[nosep,itemsep=2pt,topsep=4pt]
    \item Review all the code sets carefully, identifying common themes and unique concepts.
    \item Combine similar codes across different interviews, choosing the most descriptive and clear wording.
    \item Generalize codes when appropriate to capture broader themes that appear across multiple interviews.
    \item Retain unique codes that represent important themes specific to individual interviews.
    \item Ensure that the final set of codes is balanced, covering all major themes present in the original code sets.
    \item Aim for clarity and conciseness in your final codes, typically using one to three words per code.
    \item[\textcolor{optional}{7.}] \optionalinline{Provide a brief description (up to 20 words) for each code to clarify its meaning and application that differentiates it from other codes.}
\end{enumerate}

\vspace{5pt}

Before providing your final list of codes, use the <scratchpad> to think through your process:\\

<scratchpad>
\begin{enumerate}[nosep,itemsep=2pt,topsep=4pt]
    \item Identify recurring codes and themes across all interviews
    \item Note any unique codes that represent important individual perspectives
    \item Consider how to merge similar codes without losing nuance
    \item Reflect on potential broader categories that could encompass multiple codes
    \item Ensure all major themes from the original code sets are represented
\end{enumerate}
</scratchpad>\\

Now, please provide your comprehensive set of codes based on all the interviews. Present your codes in the following format:\\

<comprehensive\_codes>
\begin{itemize}[nosep,itemsep=2pt,topsep=4pt]
    \item Code 1 \optionalinline{| Description that explains the meaning and context of Code 1 in up to 20 words}
    \item Code 2 \optionalinline{| Description that explains the meaning and context of Code 2 in up to 20 words}
    \item Code 3 \optionalinline{| Description that explains the meaning and context of Code 3 in up to 20 words}
    \item[] ...
\end{itemize}
</comprehensive\_codes>\\

Remember, your final set should have no more than \param{max\_codes} codes. Ensure that these codes are relevant, clear, and useful for further analysis in an ethnographic study. Do not write content outside <scratchpad> or <comprehensive\_codes>.

\promptsep

\paramheading

\begin{paramlist}
    \item \param{interview\_codes}: The sets of codes from multiple interviews
    \item \param{context}: Optional additional context to consider
    \item \param{max\_codes}: Maximum number of codes to present in the final set
\end{paramlist}
\end{promptbox}
\end{figure*}

\begin{figure*}
\begin{promptbox}[title={Code Application}]

\small

You are an AI assistant tasked with analyzing an ethnographic interview and extracting relevant parts that correspond to a specific code from a given taxonomy. Follow these instructions carefully:\\

1. First, you will be presented with the full text of an interview:\\

\noindent
<interview>\\
\param{interview}\\
</interview>\\

2. Next, you will be given a taxonomy of codes \optionalinline{with its differentiating descriptions}:\\

\noindent
<taxonomy>\\
\param{set\_of\_codes}\\
</taxonomy>\\

3. You will be focusing on one specific code from this taxonomy:\\

\noindent
<code>\\
\param{specific\_code}\\
</code>\\

4. Your task is to carefully read through the interview text and identify parts that are most important or salient in relation to the specified code. These parts should justify assigning the code to those sections of the interview.\\

5. When you find relevant parts, list them in the following format:
\begin{itemize}[nosep,itemsep=2pt,topsep=4pt]
    \item \texttt{- <part>exact text from the interview</part>}
    \item \texttt{- <part>another exact text from the interview</part>}
    \item (Continue this format for all relevant parts you find)
\end{itemize}
\vspace{6pt}
\textbf{Important notes:}
\begin{itemize}[nosep,itemsep=2pt,topsep=4pt]
    \item Do not change the content of the extracted parts in any way.
    \item Include only the most relevant and important parts. Quality is more important than quantity.
    \item Ensure that each extracted part corresponds to exactly one line from the original interview. Do not merge multiple lines or extract partial lines.
    \item Ensure that the extracted parts, when taken together, provide a clear justification for assigning the specified code.
\end{itemize}
\vspace{6pt}
6. If you cannot find any parts of the interview that are relevant to the specified code, respond with:
\begin{center}
\texttt{None}
\end{center}
\vspace{6pt}
Remember, your goal is to provide an accurate and focused analysis that helps understand how the specified code applies to this interview. Be thorough in your examination but selective in your choices of relevant parts. Present your findings without any additional commentary. Start your response with either the list of parts or ``None'' if no relevant parts are found.

\promptsep

\paramheading

\begin{paramlist}
    \item \param{interview}: The full text of the ethnographic interview
    \item \param{set\_of\_codes}: The taxonomy of codes \optionalinline{with differentiating descriptions}
    \item \param{specific\_code}: The single code from the taxonomy that you must focus on
\end{paramlist}
\end{promptbox}
\end{figure*}

\begin{figure*}
\begin{promptbox}[title={Pattern Finding}]

\small

You are an AI assistant tasked with the final stage of an automated ethnography pipeline: Pattern Finding. Your goal is to analyze coded segments from multiple interviews and generate theoretical findings based on this primary, coded data.\\

You will be presented with coded segments for a specific code found across all interviews. These segments are contained in the following variable:\\

\noindent
<coded\_segments>\\
\param{coded\_segments}\\
</coded\_segments>\\

The specific code these segments relate to is:\\

\noindent
<code>\\
\param{code}\\
</code>\\

Your task is to carefully analyze these coded segments and identify meaningful patterns, themes, or theoretical findings. Follow these guidelines:
\begin{enumerate}[nosep,itemsep=2pt,topsep=4pt]
    \item Read through all the coded segments thoroughly, paying attention to recurring ideas, contradictions, and unique perspectives.
    \item Look for connections between different segments that might reveal deeper insights or patterns.
    \item Aim to generate 3--5 significant findings or patterns. Focus on quality over quantity.
    \item Prioritize non-trivial findings that go beyond surface-level observations.
    \item Each finding should be supported by evidence from multiple coded segments when possible.
\end{enumerate}

\vspace{6pt}

Before presenting your final findings, use the <scratchpad> to think through your analysis:\\

<scratchpad>
\begin{enumerate}[nosep,itemsep=2pt,topsep=4pt]
    \item Identify recurring themes or ideas across the coded segments
    \item Note any contradictions or divergent perspectives
    \item Consider how these segments relate to the specific code and the broader context of the study
    \item Reflect on potential deeper meanings or implications of the data
    \item Formulate initial ideas for findings or patterns
\end{enumerate}
</scratchpad>\\

Now, present your findings in the following format:\\

<findings>
\begin{enumerate}[nosep,itemsep=2pt,topsep=4pt]
    \item \textbf{Brief title of finding}
    \begin{sloppypar}
    [Detailed explanation of the finding, including supporting evidence from the coded segments]
    \end{sloppypar}

    \item \textbf{Brief title of finding}
    \begin{sloppypar}
    [Detailed explanation of the finding, including supporting evidence from the coded segments]
    \end{sloppypar}
    \item[3.] [Continue this format for all findings]
\end{enumerate}
</findings>\\

Remember to focus on generating insightful, non-trivial findings that contribute to a deeper understanding of the research topic. Ensure that your findings are well-supported by the data and relevant to the specific code and overall research context.

\promptsep

\paramheading

\begin{paramlist}
    \item \param{coded\_segments}: The coded segments from multiple interviews that relate to the specific code
    \item \param{code}: The code under analysis for which the segments have been collected
\end{paramlist}
\end{promptbox}
\end{figure*}

\FloatBarrier

\onecolumn

\section{Detailed Evaluation Results}

\subsection{Semantic Relatedness of Codebooks}

\begin{table}[h]
\centering
\begin{tabular}{l|c|ccccc|c}
\toprule
& \# Rel. & \multicolumn{5}{c|}{Distribution of Relationships} &  \\
& $N$ & $M$ & $C$ & $P$ & $U$ & $\tau_{sem}$ & Mean $\tau_{sem}$ \\
\midrule
\multicolumn{8}{l}{\textit{Coder A -- Coder B}} \\
Evaluator 1 & 28 & 0.154 & 0.352 & 0.185 & 0.308 & 0.493 & \multirow{3}{*}{0.584} \\
Evaluator 2 & 49 & 0.154 & 0.386 & 0.445 & 0.015 & 0.647 & \\
Evaluator 3 & 47 & 0.339 & 0.301 & 0.122 & 0.238 & 0.611 & \\
\hline
\multicolumn{8}{l}{\textit{Coder A -- AI}} \\
Evaluator 1 & 35 & 0.191 & 0.396 & 0.191 & 0.223 & 0.563 & \multirow{3}{*}{0.638} \\
Evaluator 2 & 45 & 0.064 & 0.522 & 0.382 & 0.032 & 0.620 & \\
Evaluator 3 & 101 & 0.364 & 0.523 & 0.000 & 0.113 & 0.730 & \\
\hline
\multicolumn{8}{l}{\textit{Coder B -- AI}} \\
Evaluator 1 & 36 & 0.152 & 0.517 & 0.136 & 0.195 & 0.582 & \multirow{3}{*}{0.545} \\
Evaluator 2 & 72 & 0.030 & 0.688 & 0.222 & 0.060 & 0.623 & \\
Evaluator 3 & 36 & 0.061 & 0.515 & 0.016 & 0.409 & 0.429 & \\
\bottomrule
\end{tabular}
\caption{Relationship distributions between codebooks from human coders and AICoE, as evaluated by annotators}
\label{tab:evaluator-scores-detailed}
\end{table}

\subsection{Relevance Scores of Code Assignments}

\begin{table*}[h]
\centering
\begin{subtable}[t]{0.49\textwidth}
    \centering
    \begin{tabular}{llcc}
        \toprule
        & Int. ID & Human & AI \\
        \midrule
        Evaluator 1 & 1 & 0.926 & 0.960 \\
                    & 2 & 0.984 & 0.967 \\
                    & 3 & 0.994 & 0.992 \\
        \midrule
        Evaluator 2 & 1 & 0.759 & 0.854 \\
                    & 2 & 0.875 & 0.797 \\
                    & 3 & 0.872 & 0.671 \\
        \midrule
        Evaluator 3 & 1 & 0.519 & 0.510 \\
                    & 2 & 0.661 & 0.625 \\
                    & 3 & 0.667 & 0.461 \\
        \midrule
        \multicolumn{2}{l}{Overall Average} & \textbf{0.806} & \textbf{0.760} \\
        \bottomrule
    \end{tabular}
    \caption{Scores for the CVDQuoading dataset}
    \label{tab:evaluation-metrics-cvd}
\end{subtable}
\hfill
\begin{subtable}[t]{0.49\textwidth}
    \centering
    \begin{tabular}{llcc}
        \toprule
        & Int. ID & Human & AI \\
        \midrule
        Evaluator 1 & 1 & 0.685 & 0.551 \\
                    & 2 & 0.881 & 0.643 \\
                    & 3 & 0.966 & 0.935 \\
        \midrule
        Evaluator 2 & 1 & 0.849 & 0.721 \\
                    & 2 & 0.944 & 0.599 \\
                    & 3 & 0.896 & 0.673 \\
        \midrule
        Evaluator 3 & 1 & 0.542 & 0.389 \\
                    & 2 & 0.457 & 0.224 \\
                    & 3 & 0.444 & 0.263 \\
        \midrule
        \multicolumn{2}{l}{Overall Average} & \textbf{0.740} & \textbf{0.560} \\
        \bottomrule
    \end{tabular}
    \caption{Scores for the HiAICS dataset}
    \label{tab:evaluation-metrics-ai}
\end{subtable}
\caption{Relevant code assignments from human and AI coders for each interview and evaluator}
\label{tab:rel-code-assignments-detailed}
\end{table*}

\subsection{Evaluation of Theoretical Findings}

\subsubsection{Correlation Coefficients}
\label{sec:corr_coff}

\begin{minipage}[t]{\textwidth}
\centering
\label{tab:correlation_coefficients}
\begin{tabular}{lccc}
\toprule
Criterion &   E1-E2 &   E1-E3 &  E2-E3 \\
\midrule
Grounding & -0.0430 &  0.0269 & 0.6471 \\
Relevance &  0.0064 &  0.0603 & 0.1194 \\
  Insight &  0.0846 & -0.0384 & 0.2478 \\
\bottomrule
\end{tabular}
\captionof{table}{Correlation Coefficients between Evaluators for Each Criterion}
\end{minipage}

\FloatBarrier

\newpage

\subsubsection{Quality of Theoretical Findings}

\begin{tiny}
\setlength{\tabcolsep}{2pt}
\renewcommand{\arraystretch}{1.1}

\begin{minipage}[t]{0.48\textwidth}
\centering
\begin{tabular}{lccc: c |ccc: c |ccc: c |c}
\noalign{\vskip 5mm}
\toprule
\textbf{Code} & \multicolumn{3}{c:}{\textbf{Grounding}} & \textbf{Avg} & \multicolumn{3}{c:}{\textbf{Relevance}} & \textbf{Avg} & \multicolumn{3}{c:}{\textbf{Insight}} & \textbf{Avg} & \textbf{Avg} \\
 & \textbf{E1} & \textbf{E2} & \textbf{E3} &  & \textbf{E1} & \textbf{E2} & \textbf{E3} &  & \textbf{E1} & \textbf{E2} & \textbf{E3} &  &  \\
\midrule
\multirow{5}{*}{\parbox{1cm}{\raggedright AI Critique}} &  5 &  4 &  5 &  4.67 &  4 &  4 &  3 &  3.67 &  3 &  3 &  3 &  3.00 &  3.78 \\
&  4 &  5 &  4 &  4.33 &  4 &  3 &  4 &  3.67 &  4 &  4 &  2 &  3.33 &  3.78 \\
& \cellcolor{gray!20} 4 & \cellcolor{gray!20} 4 & \cellcolor{gray!20} 5 & \cellcolor{gray!20} 4.33 & \cellcolor{gray!20} 4 & \cellcolor{gray!20} 5 & \cellcolor{gray!20} 4 & \cellcolor{gray!20} 4.33 & \cellcolor{gray!20} 3 & \cellcolor{gray!20} 5 & \cellcolor{gray!20} 4 & \cellcolor{gray!20} 4.00 & \cellcolor{gray!20} 4.22 \\
& \cellcolor{gray!20} 4 & \cellcolor{gray!20} 4 & \cellcolor{gray!20} 5 & \cellcolor{gray!20} 4.33 & \cellcolor{gray!20} 4 & \cellcolor{gray!20} 4 & \cellcolor{gray!20} 5 & \cellcolor{gray!20} 4.33 & \cellcolor{gray!20} 3 & \cellcolor{gray!20} 4 & \cellcolor{gray!20} 5 & \cellcolor{gray!20} 4.00 & \cellcolor{gray!20} 4.22 \\
&  3 &  2 &  1 &  2.00 &  4 &  2 &  4 &  3.33 &  4 &  3 &  4 &  3.67 &  3.00 \\
\hline
\multirow{5}{*}{\parbox{1cm}{\raggedright AI for Science}} &  4 &  4 &  3 &  3.67 &  4 &  5 &  4 &  4.33 &  4 &  3 &  4 &  3.67 &  3.89 \\
&  4 &  4 &  2 &  3.33 &  4 &  5 &  3 &  4.00 &  4 &  3 &  1 &  2.67 &  3.33 \\
& \cellcolor{gray!20} 4 & \cellcolor{gray!20} 5 & \cellcolor{gray!20} 5 & \cellcolor{gray!20} 4.67 & \cellcolor{gray!20} 3 & \cellcolor{gray!20} 5 & \cellcolor{gray!20} 4 & \cellcolor{gray!20} 4.00 & \cellcolor{gray!20} 4 & \cellcolor{gray!20} 3 & \cellcolor{gray!20} 3 & \cellcolor{gray!20} 3.33 & \cellcolor{gray!20} 4.00 \\
&  4 &  3 &  5 &  4.00 &  4 &  5 &  4 &  4.33 &  4 &  3 &  3 &  3.33 &  3.89 \\
&  3 &  4 &  4 &  3.67 &  4 &  5 &  4 &  4.33 &  4 &  4 &  3 &  3.67 &  3.89 \\
\hline
\multirow{5}{*}{\parbox{1cm}{\raggedright Algorithm}} &  3 &  4 &  5 &  4.00 &  3 &  4 &  3 &  3.33 &  3 &  3 &  2 &  2.67 &  3.33 \\
&  4 &  4 &  3 &  3.67 &  4 &  4 &  4 &  4.00 &  4 &  3 &  3 &  3.33 &  3.67 \\
&  4 &  3 &  3 &  3.33 &  4 &  3 &  4 &  3.67 &  4 &  3 &  4 &  3.67 &  3.56 \\
&  4 &  4 &  3 &  3.67 &  4 &  4 &  3 &  3.67 &  3 &  4 &  4 &  3.67 &  3.67 \\
&  4 &  3 &  3 &  3.33 &  4 &  4 &  4 &  4.00 &  4 &  3 &  4 &  3.67 &  3.67 \\
\hline
\multirow{5}{*}{\parbox{1cm}{\raggedright Algorithmic Biases}} & \cellcolor{gray!20} 4 & \cellcolor{gray!20} 5 & \cellcolor{gray!20} 5 & \cellcolor{gray!20} 4.67 & \cellcolor{gray!20} 4 & \cellcolor{gray!20} 5 & \cellcolor{gray!20} 4 & \cellcolor{gray!20} 4.33 & \cellcolor{gray!20} 4 & \cellcolor{gray!20} 5 & \cellcolor{gray!20} 3 & \cellcolor{gray!20} 4.00 & \cellcolor{gray!20} 4.33 \\
& \cellcolor{gray!20} 4 & \cellcolor{gray!20} 3 & \cellcolor{gray!20} 5 & \cellcolor{gray!20} 4.00 & \cellcolor{gray!20} 4 & \cellcolor{gray!20} 5 & \cellcolor{gray!20} 4 & \cellcolor{gray!20} 4.33 & \cellcolor{gray!20} 3 & \cellcolor{gray!20} 4 & \cellcolor{gray!20} 4 & \cellcolor{gray!20} 3.67 & \cellcolor{gray!20} 4.00 \\
&  4 &  3 &  3 &  3.33 &  5 &  5 &  4 &  4.67 &  4 &  4 &  3 &  3.67 &  3.89 \\
&  4 &  4 &  4 &  4.00 &  4 &  3 &  3 &  3.33 &  4 &  3 &  3 &  3.33 &  3.55 \\
&  4 &  3 &  3 &  3.33 &  4 &  3 &  3 &  3.33 &  4 &  4 &  4 &  4.00 &  3.55 \\
\hline
\multirow{5}{*}{\parbox{1cm}{\raggedright Autonomy \& Agency}} &  3 &  3 &  4 &  3.33 &  4 &  2 &  3 &  3.00 &  4 &  2 &  2 &  2.67 &  3.00 \\
&  4 &  3 &  3 &  3.33 &  4 &  2 &  3 &  3.00 &  4 &  2 &  3 &  3.00 &  3.11 \\
&  4 &  2 &  3 &  3.00 &  4 &  2 &  4 &  3.33 &  4 &  2 &  4 &  3.33 &  3.22 \\
&  3 &  2 &  3 &  2.67 &  3 &  3 &  3 &  3.00 &  3 &  2 &  2 &  2.33 &  2.67 \\
&  4 &  4 &  3 &  3.67 &  4 &  4 &  4 &  4.00 &  4 &  3 &  2 &  3.00 &  3.56 \\
\hline
\multirow{5}{*}{\parbox{1cm}{\raggedright Biographical Context}} &  4 &  5 &  4 &  4.33 &  3 &  4 &  4 &  3.67 &  3 &  4 &  4 &  3.67 &  3.89 \\
&  4 &  4 &  4 &  4.00 &  4 &  3 &  5 &  4.00 &  4 &  4 &  3 &  3.67 &  3.89 \\
&  3 &  3 &  3 &  3.00 &  3 &  4 &  4 &  3.67 &  3 &  3 &  3 &  3.00 &  3.22 \\
&  3 &  4 &  3 &  3.33 &  4 &  3 &  4 &  3.67 &  3 &  3 &  3 &  3.00 &  3.33 \\
&  4 &  3 &  3 &  3.33 &  3 &  3 &  4 &  3.33 &  3 &  3 &  4 &  3.33 &  3.33 \\
\hline
\multirow{4}{*}{\parbox{1cm}{\raggedright Black Box}} &  4 &  3 &  4 &  3.67 &  4 &  4 &  4 &  4.00 &  3 &  3 &  2 &  2.67 &  3.45 \\
&  4 &  2 &  3 &  3.00 &  4 &  3 &  4 &  3.67 &  4 &  4 &  3 &  3.67 &  3.45 \\
&  4 &  2 &  3 &  3.00 &  4 &  3 &  4 &  3.67 &  3 &  2 &  2 &  2.33 &  3.00 \\
&  2 &  2 &  3 &  2.33 &  4 &  4 &  4 &  4.00 &  3 &  3 &  4 &  3.33 &  3.22 \\
\hline
\multirow{4}{*}{\parbox{1cm}{\raggedright Data}} &  4 &  4 &  4 &  4.00 &  3 &  3 &  4 &  3.33 &  2 &  3 &  3 &  2.67 &  3.33 \\
&  4 &  3 &  3 &  3.33 &  4 &  4 &  5 &  4.33 &  4 &  4 &  3 &  3.67 &  3.78 \\
&  4 &  3 &  3 &  3.33 &  3 &  4 &  4 &  3.67 &  2 &  4 &  5 &  3.67 &  3.56 \\
&  3 &  4 &  4 &  3.67 &  3 &  4 &  4 &  3.67 &  3 &  3 &  3 &  3.00 &  3.45 \\
\hline
\multirow{5}{*}{\parbox{1cm}{\raggedright Epistemic and Infrastructural Media}} &  3 &  5 &  5 &  4.33 &  3 &  4 &  4 &  3.67 &  3 &  5 &  3 &  3.67 &  3.89 \\
&  3 &  4 &  3 &  3.33 &  3 &  4 &  5 &  4.00 &  2 &  3 &  4 &  3.00 &  3.44 \\
&  4 &  3 &  3 &  3.33 &  4 &  4 &  5 &  4.33 &  3 &  4 &  4 &  3.67 &  3.78 \\
&  3 &  3 &  4 &  3.33 &  3 &  4 &  4 &  3.67 &  3 &  3 &  3 &  3.00 &  3.33 \\
&  3 &  2 &  2 &  2.33 &  3 &  4 &  5 &  4.00 &  3 &  3 &  5 &  3.67 &  3.33 \\
\hline
\multirow{5}{*}{\parbox{1cm}{\raggedright Expert Systems}} & \cellcolor{gray!20} 4 & \cellcolor{gray!20} 4 & \cellcolor{gray!20} 4 & \cellcolor{gray!20} 4.00 & \cellcolor{gray!20} 4 & \cellcolor{gray!20} 4 & \cellcolor{gray!20} 5 & \cellcolor{gray!20} 4.33 & \cellcolor{gray!20} 4 & \cellcolor{gray!20} 4 & \cellcolor{gray!20} 3 & \cellcolor{gray!20} 3.67 & \cellcolor{gray!20} 4.00 \\
&  3 &  3 &  4 &  3.33 &  4 &  5 &  4 &  4.33 &  3 &  3 &  3 &  3.00 &  3.55 \\
&  3 &  3 &  4 &  3.33 &  3 &  5 &  3 &  3.67 &  3 &  3 &  3 &  3.00 &  3.33 \\
&  4 &  3 &  3 &  3.33 &  4 &  4 &  5 &  4.33 &  4 &  3 &  4 &  3.67 &  3.78 \\
&  4 &  3 &  3 &  3.33 &  3 &  4 &  4 &  3.67 &  3 &  3 &  4 &  3.33 &  3.44 \\
\hline
\multirow{5}{*}{\parbox{1cm}{\raggedright Expertise Competence}} &  3 &  4 &  4 &  3.67 &  4 &  5 &  3 &  4.00 &  4 &  3 &  2 &  3.00 &  3.56 \\
&  4 &  3 &  4 &  3.67 &  4 &  4 &  4 &  4.00 &  4 &  2 &  3 &  3.00 &  3.56 \\
&  4 &  3 &  4 &  3.67 &  4 &  4 &  4 &  4.00 &  4 &  3 &  4 &  3.67 &  3.78 \\
&  4 &  1 &  3 &  2.67 &  4 &  3 &  3 &  3.33 &  3 &  2 &  3 &  2.67 &  2.89 \\
&  4 &  3 &  4 &  3.67 &  4 &  4 &  4 &  4.00 &  4 &  3 &  3 &  3.33 &  3.67 \\
\hline
\multirow{5}{*}{\parbox{1cm}{\raggedright Facial Recognition}} &  5 &  3 &  4 &  4.00 &  4 &  2 &  4 &  3.33 &  4 &  4 &  4 &  4.00 &  3.78 \\
&  4 &  3 &  2 &  3.00 &  4 &  4 &  4 &  4.00 &  4 &  3 &  3 &  3.33 &  3.44 \\
&  4 &  2 &  2 &  2.67 &  4 &  3 &  4 &  3.67 &  3 &  3 &  3 &  3.00 &  3.11 \\
&  3 &  2 &  2 &  2.33 &  3 &  4 &  3 &  3.33 &  3 &  2 &  2 &  2.33 &  2.66 \\
&  4 &  1 &  2 &  2.33 &  3 &  3 &  4 &  3.33 &  3 &  3 &  4 &  3.33 &  3.00 \\
\hline
\multirow{5}{*}{\parbox{1cm}{\raggedright First Encounters with AI}} & \cellcolor{gray!20} 4 & \cellcolor{gray!20} 5 & \cellcolor{gray!20} 4 & \cellcolor{gray!20} 4.33 & \cellcolor{gray!20} 4 & \cellcolor{gray!20} 4 & \cellcolor{gray!20} 4 & \cellcolor{gray!20} 4.00 & \cellcolor{gray!20} 4 & \cellcolor{gray!20} 4 & \cellcolor{gray!20} 4 & \cellcolor{gray!20} 4.00 & \cellcolor{gray!20} 4.11 \\
&  3 &  3 &  3 &  3.00 &  3 &  4 &  4 &  3.67 &  3 &  3 &  3 &  3.00 &  3.22 \\
&  2 &  4 &  4 &  3.33 &  2 &  4 &  4 &  3.33 &  2 &  2 &  3 &  2.33 &  3.00 \\
&  3 &  3 &  3 &  3.00 &  3 &  4 &  4 &  3.67 &  3 &  3 &  4 &  3.33 &  3.33 \\
&  4 &  3 &  3 &  3.33 &  4 &  4 &  4 &  4.00 &  4 &  4 &  3 &  3.67 &  3.67 \\
\hline
\multirow{5}{*}{\parbox{1cm}{\raggedright Format}} &  3 &  2 &  4 &  3.00 &  3 &  3 &  3 &  3.00 &  3 &  3 &  2 &  2.67 &  2.89 \\
&  3 &  2 &  3 &  2.67 &  4 &  2 &  3 &  3.00 &  3 &  2 &  4 &  3.00 &  2.89 \\
&  4 &  3 &  3 &  3.33 &  4 &  3 &  4 &  3.67 &  4 &  4 &  4 &  4.00 &  3.67 \\
&  4 &  2 &  3 &  3.00 &  4 &  3 &  4 &  3.67 &  3 &  3 &  4 &  3.33 &  3.33 \\
&  3 &  2 &  2 &  2.33 &  4 &  3 &  3 &  3.33 &  4 &  3 &  4 &  3.67 &  3.11 \\
\hline
\multirow{5}{*}{\parbox{1cm}{\raggedright Generative AI}} &  2 &  5 &  4 &  3.67 &  3 &  4 &  4 &  3.67 &  3 &  4 &  3 &  3.33 &  3.56 \\
&  4 &  3 &  3 &  3.33 &  3 &  4 &  3 &  3.33 &  3 &  3 &  4 &  3.33 &  3.33 \\
&  4 &  4 &  4 &  4.00 &  3 &  3 &  4 &  3.33 &  3 &  2 &  3 &  2.67 &  3.33 \\
&  3 &  2 &  1 &  2.00 &  3 &  4 &  3 &  3.33 &  3 &  3 &  2 &  2.67 &  2.67 \\
&  4 &  3 &  3 &  3.33 &  4 &  5 &  4 &  4.33 &  4 &  3 &  4 &  3.67 &  3.78 \\
\hline
\multirow{5}{*}{\parbox{1cm}{\raggedright Historical Perspectives on AI, ML, ANN}} &  3 &  3 &  4 &  3.33 &  3 &  4 &  4 &  3.67 &  3 &  4 &  3 &  3.33 &  3.44 \\
&  3 &  3 &  4 &  3.33 &  3 &  2 &  4 &  3.00 &  3 &  2 &  3 &  2.67 &  3.00 \\
&  3 &  3 &  3 &  3.00 &  3 &  3 &  4 &  3.33 &  3 &  2 &  3 &  2.67 &  3.00 \\
&  3 &  3 &  3 &  3.00 &  3 &  3 &  4 &  3.33 &  3 &  3 &  2 &  2.67 &  3.00 \\
&  3 &  3 &  3 &  3.00 &  3 &  4 &  4 &  3.67 &  3 &  3 &  3 &  3.00 &  3.22 \\
\bottomrule
\end{tabular}
\end{minipage}
\hfill
\begin{minipage}[t]{0.48\textwidth}
\centering
\begin{tabular}{lccc: c |ccc: c |ccc: c |c}
\noalign{\vskip -5.8mm}
\toprule
\textbf{Code} & \multicolumn{3}{c:}{\textbf{Grounding}} & \textbf{Avg} & \multicolumn{3}{c:}{\textbf{Relevance}} & \textbf{Avg} & \multicolumn{3}{c:}{\textbf{Insight}} & \textbf{Avg} & \textbf{Avg} \\
 & \textbf{E1} & \textbf{E2} & \textbf{E3} &  & \textbf{E1} & \textbf{E2} & \textbf{E3} &  & \textbf{E1} & \textbf{E2} & \textbf{E3} & & \\
\midrule
\multirow{5}{*}{\parbox{1cm}{\raggedright Images}} &  5 &  4 &  5 &  4.67 &  4 &  4 &  3 &  3.67 &  3 &  3 &  3 &  3.00 &  3.78 \\
&  4 &  5 &  4 &  4.33 &  4 &  3 &  4 &  3.67 &  4 &  4 &  2 &  3.33 &  3.78 \\
& \cellcolor{gray!20} 4 & \cellcolor{gray!20} 4 & \cellcolor{gray!20} 5 & \cellcolor{gray!20} 4.33 & \cellcolor{gray!20} 4 & \cellcolor{gray!20} 5 & \cellcolor{gray!20} 4 & \cellcolor{gray!20} 4.33 & \cellcolor{gray!20} 3 & \cellcolor{gray!20} 5 & \cellcolor{gray!20} 4 & \cellcolor{gray!20} 4.00 & \cellcolor{gray!20} 4.22 \\
& \cellcolor{gray!20} 4 & \cellcolor{gray!20} 4 & \cellcolor{gray!20} 5 & \cellcolor{gray!20} 4.33 & \cellcolor{gray!20} 4 & \cellcolor{gray!20} 4 & \cellcolor{gray!20} 5 & \cellcolor{gray!20} 4.33 & \cellcolor{gray!20} 3 & \cellcolor{gray!20} 4 & \cellcolor{gray!20} 5 & \cellcolor{gray!20} 4.00 & \cellcolor{gray!20} 4.22 \\
&  3 &  2 &  1 &  2.00 &  4 &  2 &  4 &  3.33 &  4 &  3 &  4 &  3.67 &  3.00 \\
\hline
\multirow{5}{*}{\parbox{1cm}{\raggedright Institutions}} &  4 &  4 &  3 &  3.67 &  4 &  5 &  4 &  4.33 &  4 &  3 &  4 &  3.67 &  3.89 \\
&  4 &  4 &  2 &  3.33 &  4 &  5 &  3 &  4.00 &  4 &  3 &  1 &  2.67 &  3.33 \\
& \cellcolor{gray!20} 4 & \cellcolor{gray!20} 5 & \cellcolor{gray!20} 5 & \cellcolor{gray!20} 4.67 & \cellcolor{gray!20} 3 & \cellcolor{gray!20} 5 & \cellcolor{gray!20} 4 & \cellcolor{gray!20} 4.00 & \cellcolor{gray!20} 4 & \cellcolor{gray!20} 3 & \cellcolor{gray!20} 3 & \cellcolor{gray!20} 3.33 & \cellcolor{gray!20} 4.00 \\
&  4 &  3 &  5 &  4.00 &  4 &  5 &  4 &  4.33 &  4 &  3 &  3 &  3.33 &  3.89 \\
&  3 &  4 &  4 &  3.67 &  4 &  5 &  4 &  4.33 &  4 &  4 &  3 &  3.67 &  3.89 \\
\hline
\multirow{5}{*}{\parbox{1cm}{\raggedright Machine Learning, ANN \& DL}} &  3 &  4 &  5 &  4.00 &  3 &  4 &  3 &  3.33 &  3 &  3 &  2 &  2.67 &  3.33 \\
&  4 &  4 &  3 &  3.67 &  4 &  4 &  4 &  4.00 &  4 &  3 &  3 &  3.33 &  3.67 \\
&  4 &  3 &  3 &  3.33 &  4 &  3 &  4 &  3.67 &  4 &  3 &  4 &  3.67 &  3.56 \\
&  4 &  4 &  3 &  3.67 &  4 &  4 &  3 &  3.67 &  3 &  4 &  4 &  3.67 &  3.67 \\
&  4 &  3 &  3 &  3.33 &  4 &  4 &  4 &  4.00 &  4 &  3 &  4 &  3.67 &  3.67 \\
\hline
\multirow{5}{*}{\parbox{1cm}{\raggedright Media Studies and Visual Culture Studies}} & \cellcolor{gray!20} 4 & \cellcolor{gray!20} 5 & \cellcolor{gray!20} 5 & \cellcolor{gray!20} 4.67 & \cellcolor{gray!20} 4 & \cellcolor{gray!20} 5 & \cellcolor{gray!20} 4 & \cellcolor{gray!20} 4.33 & \cellcolor{gray!20} 4 & \cellcolor{gray!20} 5 & \cellcolor{gray!20} 3 & \cellcolor{gray!20} 4.00 & \cellcolor{gray!20} 4.33 \\
& \cellcolor{gray!20} 4 & \cellcolor{gray!20} 3 & \cellcolor{gray!20} 5 & \cellcolor{gray!20} 4.00 & \cellcolor{gray!20} 4 & \cellcolor{gray!20} 5 & \cellcolor{gray!20} 4 & \cellcolor{gray!20} 4.33 & \cellcolor{gray!20} 3 & \cellcolor{gray!20} 4 & \cellcolor{gray!20} 4 & \cellcolor{gray!20} 3.67 & \cellcolor{gray!20} 4.00 \\
&  4 &  3 &  3 &  3.33 &  5 &  5 &  4 &  4.67 &  4 &  4 &  3 &  3.67 &  3.89 \\
&  4 &  4 &  4 &  4.00 &  4 &  3 &  3 &  3.33 &  4 &  3 &  3 &  3.33 &  3.55 \\
&  4 &  3 &  3 &  3.33 &  4 &  3 &  3 &  3.33 &  4 &  4 &  4 &  4.00 &  3.55 \\
\hline
\multirow{5}{*}{\parbox{1cm}{\raggedright Pattern Recognition}} &  3 &  3 &  4 &  3.33 &  4 &  2 &  3 &  3.00 &  4 &  2 &  2 &  2.67 &  3.00 \\
&  4 &  3 &  3 &  3.33 &  4 &  2 &  3 &  3.00 &  4 &  2 &  3 &  3.00 &  3.11 \\
&  4 &  2 &  3 &  3.00 &  4 &  2 &  4 &  3.33 &  4 &  2 &  4 &  3.33 &  3.22 \\
&  3 &  2 &  3 &  2.67 &  3 &  3 &  3 &  3.00 &  3 &  2 &  2 &  2.33 &  2.67 \\
&  4 &  4 &  3 &  3.67 &  4 &  4 &  4 &  4.00 &  4 &  3 &  2 &  3.00 &  3.56 \\
\hline
\multirow{5}{*}{\parbox{1cm}{\raggedright Political \& Economic Contexts of (Applied) AI}} &  4 &  5 &  4 &  4.33 &  3 &  4 &  4 &  3.67 &  3 &  4 &  4 &  3.67 &  3.89 \\
&  4 &  4 &  4 &  4.00 &  4 &  3 &  5 &  4.00 &  4 &  4 &  3 &  3.67 &  3.89 \\
&  3 &  3 &  3 &  3.00 &  3 &  4 &  4 &  3.67 &  3 &  3 &  3 &  3.00 &  3.22 \\
&  3 &  4 &  3 &  3.33 &  4 &  3 &  4 &  3.67 &  3 &  3 &  3 &  3.00 &  3.33 \\
&  4 &  3 &  3 &  3.33 &  3 &  3 &  4 &  3.33 &  3 &  3 &  4 &  3.33 &  3.33 \\
\hline
\multirow{4}{*}{\parbox{1cm}{\raggedright Project Description}} &  4 &  3 &  4 &  3.67 &  4 &  4 &  4 &  4.00 &  3 &  3 &  2 &  2.67 &  3.45 \\
&  4 &  2 &  3 &  3.00 &  4 &  3 &  4 &  3.67 &  4 &  4 &  3 &  3.67 &  3.45 \\
&  4 &  2 &  3 &  3.00 &  4 &  3 &  4 &  3.67 &  3 &  2 &  2 &  2.33 &  3.00 \\
&  2 &  2 &  3 &  2.33 &  4 &  4 &  4 &  4.00 &  3 &  3 &  4 &  3.33 &  3.22 \\
\hline
\multirow{4}{*}{\parbox{1cm}{\raggedright Publications}} &  4 &  4 &  4 &  4.00 &  3 &  3 &  4 &  3.33 &  2 &  3 &  3 &  2.67 &  3.33 \\
&  4 &  3 &  3 &  3.33 &  4 &  4 &  5 &  4.33 &  4 &  4 &  3 &  3.67 &  3.78 \\
&  4 &  3 &  3 &  3.33 &  3 &  4 &  4 &  3.67 &  2 &  4 &  5 &  3.67 &  3.56 \\
&  3 &  4 &  4 &  3.67 &  3 &  4 &  4 &  3.67 &  3 &  3 &  3 &  3.00 &  3.45 \\
\hline
\multirow{5}{*}{\parbox{1cm}{\raggedright Research Interest Challenges Limitations}} &  3 &  5 &  5 &  4.33 &  3 &  4 &  4 &  3.67 &  3 &  5 &  3 &  3.67 &  3.89 \\
&  3 &  4 &  3 &  3.33 &  3 &  4 &  5 &  4.00 &  2 &  3 &  4 &  3.00 &  3.44 \\
&  4 &  3 &  3 &  3.33 &  4 &  4 &  5 &  4.33 &  3 &  4 &  4 &  3.67 &  3.78 \\
&  3 &  3 &  4 &  3.33 &  3 &  4 &  4 &  3.67 &  3 &  3 &  3 &  3.00 &  3.33 \\
&  3 &  2 &  2 &  2.33 &  3 &  4 &  5 &  4.00 &  3 &  3 &  5 &  3.67 &  3.33 \\
\hline
\multirow{5}{*}{\parbox{1cm}{\raggedright Sensors \& Infrastructures \& Platforms}} & \cellcolor{gray!20} 4 & \cellcolor{gray!20} 4 & \cellcolor{gray!20} 4 & \cellcolor{gray!20} 4.00 & \cellcolor{gray!20} 4 & \cellcolor{gray!20} 4 & \cellcolor{gray!20} 5 & \cellcolor{gray!20} 4.33 & \cellcolor{gray!20} 4 & \cellcolor{gray!20} 4 & \cellcolor{gray!20} 3 & \cellcolor{gray!20} 3.67 & \cellcolor{gray!20} 4.00 \\
&  3 &  3 &  4 &  3.33 &  4 &  5 &  4 &  4.33 &  3 &  3 &  3 &  3.00 &  3.55 \\
&  3 &  3 &  4 &  3.33 &  3 &  5 &  3 &  3.67 &  3 &  3 &  3 &  3.00 &  3.33 \\
&  4 &  3 &  3 &  3.33 &  4 &  4 &  5 &  4.33 &  4 &  3 &  4 &  3.67 &  3.78 \\
&  4 &  3 &  3 &  3.33 &  3 &  4 &  4 &  3.67 &  3 &  3 &  4 &  3.33 &  3.44 \\
\hline
\multirow{5}{*}{\parbox{1cm}{\raggedright Speculations Ideologies Imaginations of AI}} &  3 &  4 &  4 &  3.67 &  4 &  5 &  3 &  4.00 &  4 &  3 &  2 &  3.00 &  3.56 \\
&  4 &  3 &  4 &  3.67 &  4 &  4 &  4 &  4.00 &  4 &  2 &  3 &  3.00 &  3.56 \\
&  4 &  3 &  4 &  3.67 &  4 &  4 &  4 &  4.00 &  4 &  3 &  4 &  3.67 &  3.78 \\
&  4 &  1 &  3 &  2.67 &  4 &  3 &  3 &  3.33 &  3 &  2 &  3 &  2.67 &  2.89 \\
&  4 &  3 &  4 &  3.67 &  4 &  4 &  4 &  4.00 &  4 &  3 &  3 &  3.33 &  3.67 \\
\hline
\multirow{5}{*}{\parbox{1cm}{\raggedright Terms \& Definitions}} &  5 &  3 &  4 &  4.00 &  4 &  2 &  4 &  3.33 &  4 &  4 &  4 &  4.00 &  3.78 \\
&  4 &  3 &  2 &  3.00 &  4 &  4 &  4 &  4.00 &  4 &  3 &  3 &  3.33 &  3.44 \\
&  4 &  2 &  2 &  2.67 &  4 &  3 &  4 &  3.67 &  3 &  3 &  3 &  3.00 &  3.11 \\
&  3 &  2 &  2 &  2.33 &  3 &  4 &  3 &  3.33 &  3 &  2 &  2 &  2.33 &  2.66 \\
&  4 &  1 &  2 &  2.33 &  3 &  3 &  4 &  3.33 &  3 &  3 &  4 &  3.33 &  3.00 \\
\hline
\multirow{5}{*}{\parbox{1cm}{\raggedright Tools \& Methods}} & \cellcolor{gray!20} 4 & \cellcolor{gray!20} 5 & \cellcolor{gray!20} 4 & \cellcolor{gray!20} 4.33 & \cellcolor{gray!20} 4 & \cellcolor{gray!20} 4 & \cellcolor{gray!20} 4 & \cellcolor{gray!20} 4.00 & \cellcolor{gray!20} 4 & \cellcolor{gray!20} 4 & \cellcolor{gray!20} 4 & \cellcolor{gray!20} 4.00 & \cellcolor{gray!20} 4.11 \\
&  3 &  3 &  3 &  3.00 &  3 &  4 &  4 &  3.67 &  3 &  3 &  3 &  3.00 &  3.22 \\
&  2 &  4 &  4 &  3.33 &  2 &  4 &  4 &  3.33 &  2 &  2 &  3 &  2.33 &  3.00 \\
&  3 &  3 &  3 &  3.00 &  3 &  4 &  4 &  3.67 &  3 &  3 &  4 &  3.33 &  3.33 \\
&  4 &  3 &  3 &  3.33 &  4 &  4 &  4 &  4.00 &  4 &  4 &  3 &  3.67 &  3.67 \\
\hline
\multirow{5}{*}{\parbox{1cm}{\raggedright Trust}} &  3 &  2 &  4 &  3.00 &  3 &  3 &  3 &  3.00 &  3 &  3 &  2 &  2.67 &  2.89 \\
&  3 &  2 &  3 &  2.67 &  4 &  2 &  3 &  3.00 &  3 &  2 &  4 &  3.00 &  2.89 \\
&  4 &  3 &  3 &  3.33 &  4 &  3 &  4 &  3.67 &  4 &  4 &  4 &  4.00 &  3.67 \\
&  4 &  2 &  3 &  3.00 &  4 &  3 &  4 &  3.67 &  3 &  3 &  4 &  3.33 &  3.33 \\
&  3 &  2 &  2 &  2.33 &  4 &  3 &  3 &  3.33 &  4 &  3 &  4 &  3.67 &  3.11 \\
\hline
\multirow{5}{*}{\parbox{1cm}{\raggedright Uses of AI for...}} &  2 &  5 &  4 &  3.67 &  3 &  4 &  4 &  3.67 &  3 &  4 &  3 &  3.33 &  3.56 \\
&  4 &  3 &  3 &  3.33 &  3 &  4 &  3 &  3.33 &  3 &  3 &  4 &  3.33 &  3.33 \\
&  4 &  4 &  4 &  4.00 &  3 &  3 &  4 &  3.33 &  3 &  2 &  3 &  2.67 &  3.33 \\
&  3 &  2 &  1 &  2.00 &  3 &  4 &  3 &  3.33 &  3 &  3 &  2 &  2.67 &  2.67 \\
&  4 &  3 &  3 &  3.33 &  4 &  5 &  4 &  4.33 &  4 &  3 &  4 &  3.67 &  3.78 \\
\hline
\textbf{Average} & 3.64 & 3.22 & 3.41 & 3.42 & 3.66 & 3.75 & 3.86 & 3.76 & 3.41 & 3.19 & 3.27 & 3.29 & 3.49 \\
\bottomrule
\end{tabular}
\end{minipage}
\captionof{table}{Evaluation results of all findings for all three evaluators and criteria}
\label{tab:detailed-findings}

\end{tiny}

\newpage

\FloatBarrier

\section{Examplary Outputs}

\subsection{Codebooks}

\begin{figure}[ht]
    \centering
    \begin{tcolorbox}[enhanced, width=\textwidth, colframe=black!80, colback=white, rounded corners]
    \small{\normalsize{\textbf{Codebook Comparison}}}
    \\[8pt]
    
    \begin{minipage}[t]{0.31\textwidth}
        \centering
        \colorbox{LightBlue!30}{\textbf{Coder 1}}
        \\[6pt]
        
        \begin{itemize}[leftmargin=*,nosep,label=\textcolor{NavyBlue}{$\bullet$}]
            \item AI Critique
            \item AI for Science
            \item Algorithm
            \item Algorithmic Biases
            \item Autonomy \& Agency
            \item Biographical Context
            \item Black Box
            \item Data
            \item Epistemic and Infrastructural of Media
            \item Expert Systems
            \item Expertise \& Competence
            \item Facial Recognition
            \item First Encounters with AI
            \item Format
            \item Generative AI
            \item Historical Perspectives on AI, ML, ANN
            \item Images
            \item Institutions
            \item Machine Learning, ANN, DL
            \item Media Studies - Bildwissenschaft - Visual Culture Studies
            \item Pattern Recognition
            \item Political \& Economic Contexts of (Applied) AI
            \item Project Description
            \item Publications
            \item Research Interest, Challenges, Limitations
            \item Sensors, Infrastructures \& Platforms
            \item Speculations, Ideologies, Imaginations of AI
            \item Terms \& Definitions
            \item Tools \& Methods
            \item Trust
            \item Uses of AI for...
        \end{itemize}
    \end{minipage}
    \hfill
    \begin{minipage}[t]{0.31\textwidth}
        \centering
        \colorbox{MediumSeaGreen!30}{\textbf{Coder 2}}
        \\[6pt]
        
        \begin{itemize}[leftmargin=*,nosep,label=\textcolor{ForestGreen}{$\bullet$}]
            \item AI Critique
            \item AI History
            \item Automation of Work
            \item Commercialization
            \item Continuities In Research
            \item Data Availability
            \item Data Practices
            \item Definition of Discipline
            \item Depiction of AI
            \item Expertise
            \item Future Areas of Research
            \item History of Climate Science
            \item History of Discipline
            \item History of Facial Recognition
            \item History of Photography
            \item History of Physics
            \item Interview Technicalities
            \item Large Language Models
            \item Limitations of AI
            \item New Questions Through AI
            \item Pattern Recognition
            \item Personal Approach To AI
            \item Philosophical Implications of AI
            \item Politics of Infrastructure
            \item Possible AI Applications
            \item Practices In Climate Science
            \item Prediction
            \item Programming Practices
            \item Recent Developments In Research
            \item Recent Personal Work
            \item Recent Publications
            \item Research Practice
            \item Rule-Based AI
            \item Ruptures Through AI
        \end{itemize}
    \end{minipage}
    \hfill
    \begin{minipage}[t]{0.31\textwidth}
        \centering
        \colorbox{Salmon!30}{\textbf{AICoE}}
        \\[6pt]
        
        \begin{itemize}[leftmargin=*,nosep,label=\textcolor{Maroon}{$\bullet$}]
            \item AI Applications
            \item AI Critique
            \item Automation
            \item Bildwissenschaft
            \item Black Box Problem
            \item Climate Science
            \item Critical Theory
            \item Data Quality
            \item Digital Literacy
            \item Epistemological Questions
            \item Epistemology
            \item Ethics
            \item Extractivism
            \item Facial Recognition
            \item Future Directions
            \item Fuzziness
            \item Generative AI
            \item Human-AI Interaction
            \item Image Manipulation
            \item Infrastructures
            \item Interdisciplinary
            \item Machine Learning
            \item Media Influence
            \item Model Limitations
            \item Neocolonialism
            \item Neural Networks
            \item Pattern Recognition
            \item Prediction Challenges
            \item Style Transfer
            \item Surveillance Capitalism
            \item Uncertainty
            \item Visual Culture
        \end{itemize}
    \end{minipage}
    
    \vspace{8pt}
    \begin{minipage}{\textwidth}
    \centering
    \footnotesize
    \end{minipage}
    \end{tcolorbox}
    \caption{Side-by-side comparison of the codebooks developed by two human coders and the AICoE system for analyzing the HiAICS data. The comparison highlights overlapping themes, distinct coding approaches, and varying emphases in categories such as technical concepts, historical perspectives, ethical considerations, and individual interviewee experiences.}
        \label{fig:codebook-comparison}
\end{figure}

\newpage

\subsection{Codebook Relations}

\begin{figure}[h]
    \centering
    \includegraphics[trim={0.7cm 0.7cm 1.55cm 2.3cm},clip,width=1.0\linewidth]{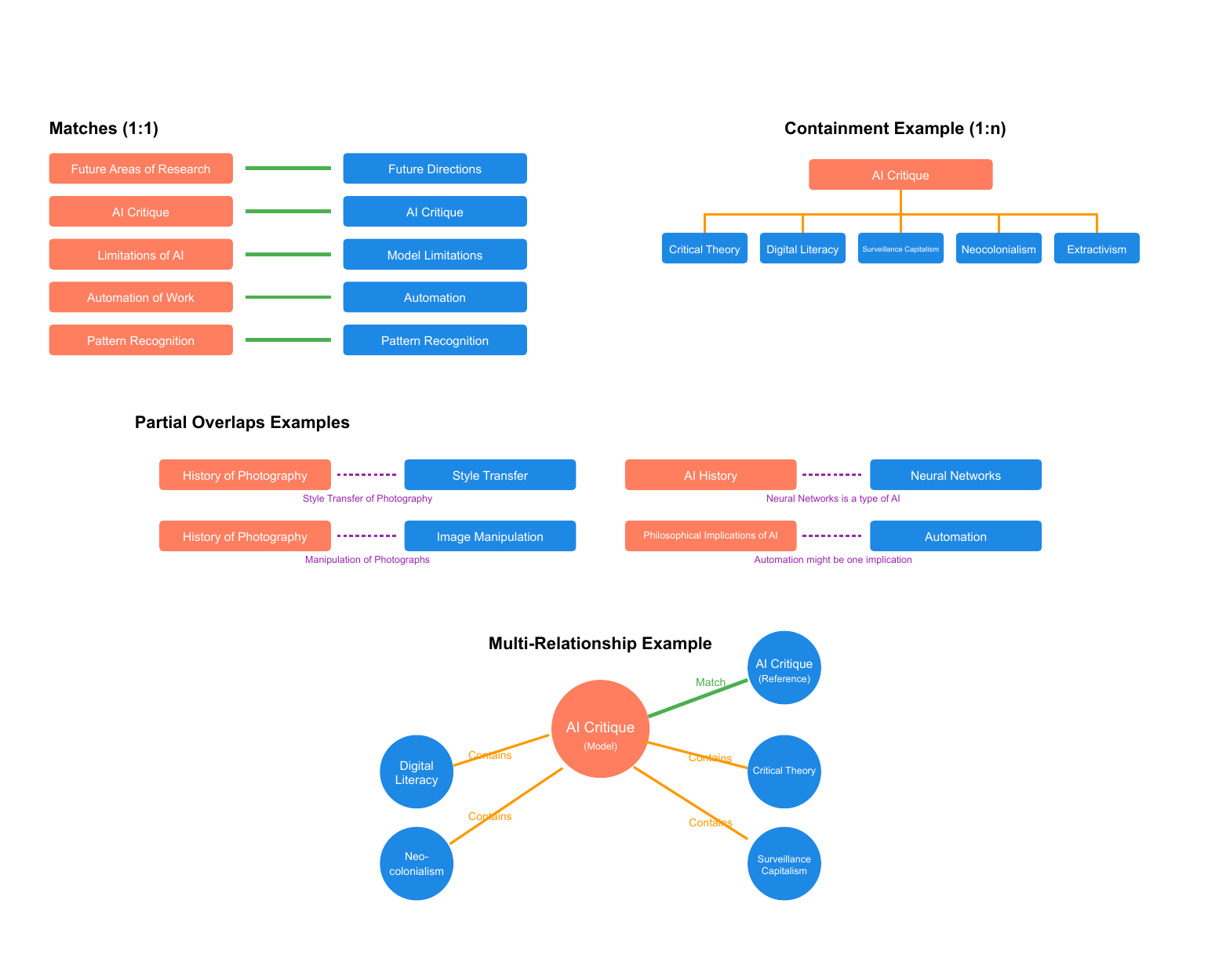}
    \caption{Examplary visualization of select relationships between codes between a human-developed codebook and the codebook of AICoE, as annotated by one of our expert annotators}
    \label{fig:codebook-relations-viz}
\end{figure}

\newpage

\subsection{Findings}

\begin{figure}[ht]
    \centering
\begin{tcolorbox}[
    colback=white,
    colframe=black,
    arc=7pt,
    boxrule=1pt,
    width=\textwidth,
    top=-2pt,
    bottom=8pt,     
    left=8pt,       
    right=8pt       
]
\begin{align*}
& \textbf{Finding 1} \hspace{2em} \text{Quality Score: } \textcolor{blue}{4.33} \\[0.5em]
& \text{\textit{Pervasiveness of Algorithmic Biases}} \\[0.5em]
& \begin{minipage}{0.95\textwidth}
    The coded segments illustrate that algorithmic biases are not limited to a specific domain but are a widespread issue affecting various applications of AI and machine learning. For instance, Speaker 0 in Interview\_\rule{1.5cm}{0.6em}\_20240905 discusses how biases can lead to incorrect predictions in climate modeling when the system encounters new, unseen data. Similarly, Speaker 0 in Interview\_\rule{1.5cm}{0.6em}\_20241016 highlights the persistent problem of bias in facial recognition technology. This pervasiveness underscores the need for a comprehensive approach to addressing biases, one that considers the unique challenges and implications of each domain.
\end{minipage} \\[1em]
& \textbf{Finding 2} \hspace{2em} \text{Quality Score: } \textcolor{blue}{4.33} \\[0.5em]
& \text{\textit{Interdisciplinary Approach to Visual Culture}} \\[0.5em]
& \begin{minipage}{0.95\textwidth}
    The coded segments suggest that combining Bildwissenschaft (focusing on the singular, autonomous image) with media studies (considering image economies and infrastructures) provides a more comprehensive understanding of AI's impact on visual culture. This is evident in Speaker 0's remark from Interview\_\rule{1.5cm}{0.6em}\_20241016, where they mention the need to combine approaches from Bildwissenschaft with media studies to deal with both the historical, singular image and the broader image ecologies. This integration is crucial for navigating the changing landscape of visual content production and analysis, especially with the advent of AI-generated images.
\end{minipage} \\[1em]
& \textbf{Finding 3} \hspace{2em} \text{Quality Score: } \textcolor{blue}{4.00} \\[0.5em]
& \text{\textit{Evolution of Expert Systems}} \\[0.5em]
& \begin{minipage}{0.95\textwidth}
    The concept of expert systems has undergone significant evolution, from being heavily reliant on rule-based systems and knowledge engineering to embracing more data-driven approaches. This shift is evident in Speaker 1's discussion from Interview\_\rule{1.5cm}{0.6em}\_20141016, where they mention, ``Today, if you want to build a similar concept, an expert system, instead of interviewing the experts, medical doctors asking them about, tell me about these symptoms and this illness and this, et cetera, you would take data, raw data.'' This evolution suggests a move towards leveraging machine learning and potentially generative AI models, as hinted at with the mention of ``generative pre-trained transformer'' in the same interview.
\end{minipage}
\end{align*}
\end{tcolorbox}
    \caption{High-quality findings generated by AI Co-Ethnographer from the HiAICS dataset, as rated by three evaluators. The Quality Score (1.00–5.00) represents the average across all evaluators and criteria.}
    \label{fig:high-quality-findings-examples}
\end{figure}

\newpage

\section{Human Evaluation Interfaces}

\begin{figure}[H]
    \centering
    \includegraphics[width=0.60\textwidth]{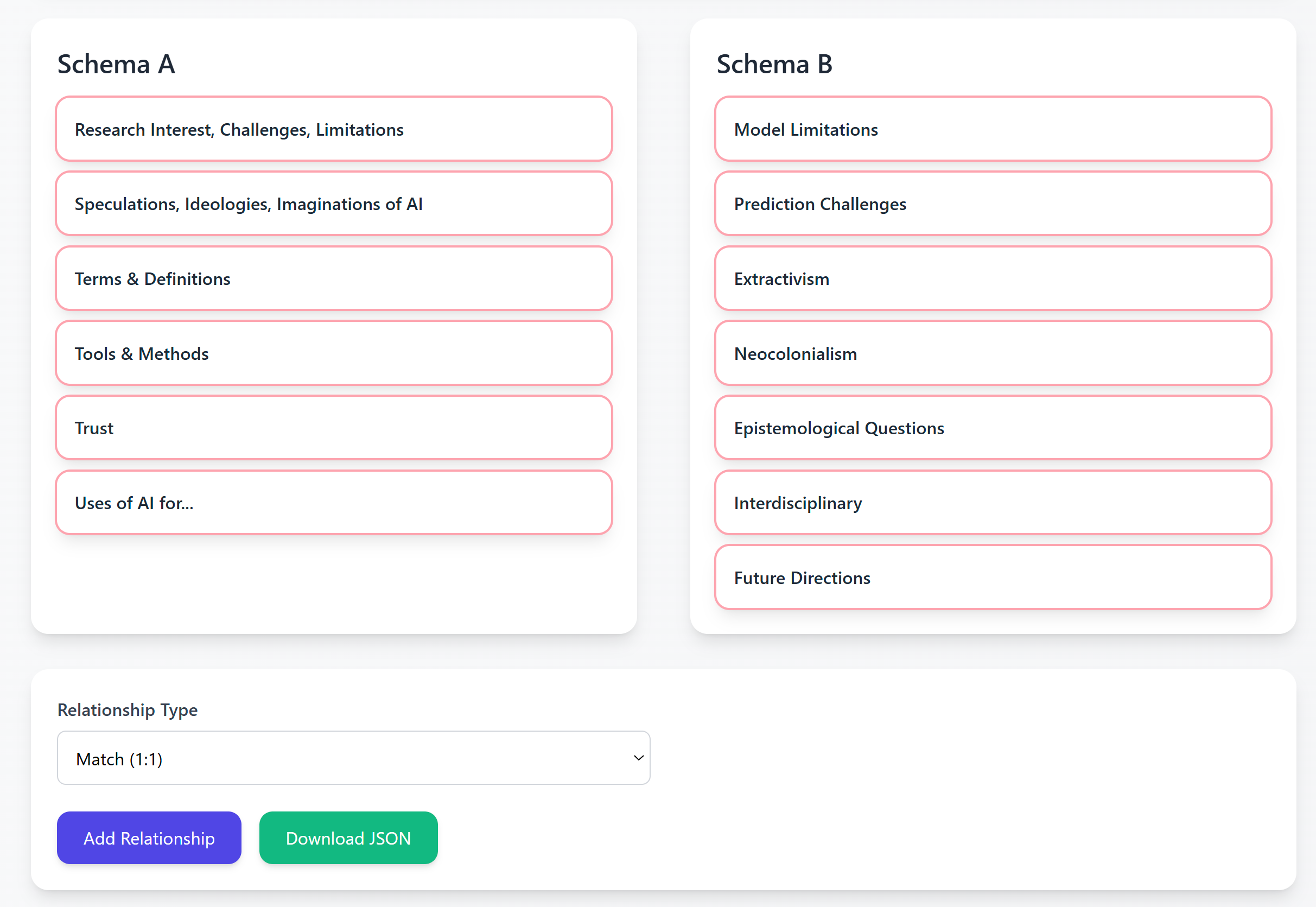}
    \caption{Evaluation interface that allows human annotators to specify the relationships between different codebooks}
    \label{fig:codebook-comparison}
\end{figure}

\begin{figure}[H]
    \centering
    \includegraphics[width=0.625\linewidth]{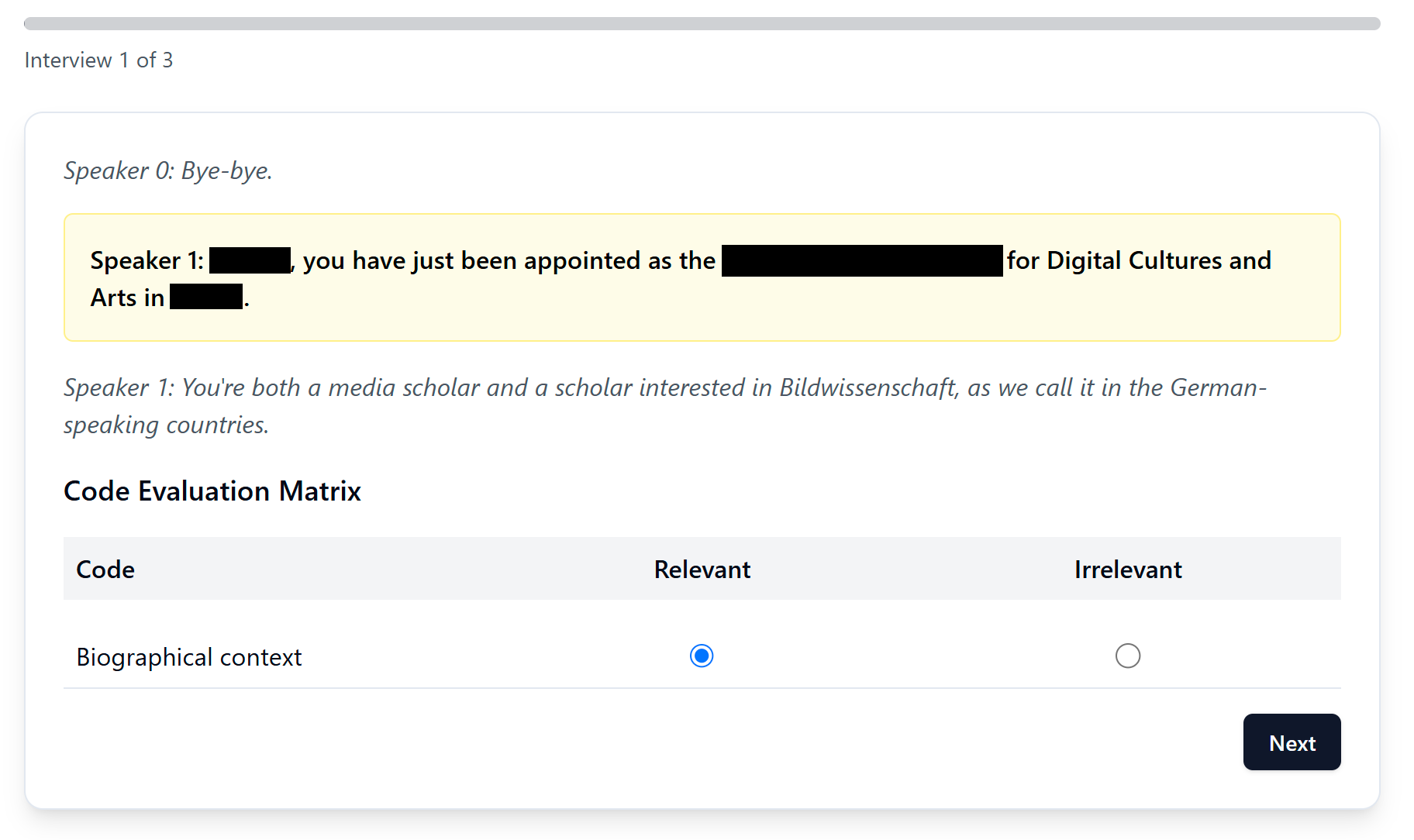}
    \caption{Evaluation interface used by human annotators to assess the relevance of code assignments}
    \label{fig:code-application-evaluation}
\end{figure}

\begin{figure}[H]
    \centering
    \includegraphics[width=0.7725\linewidth]{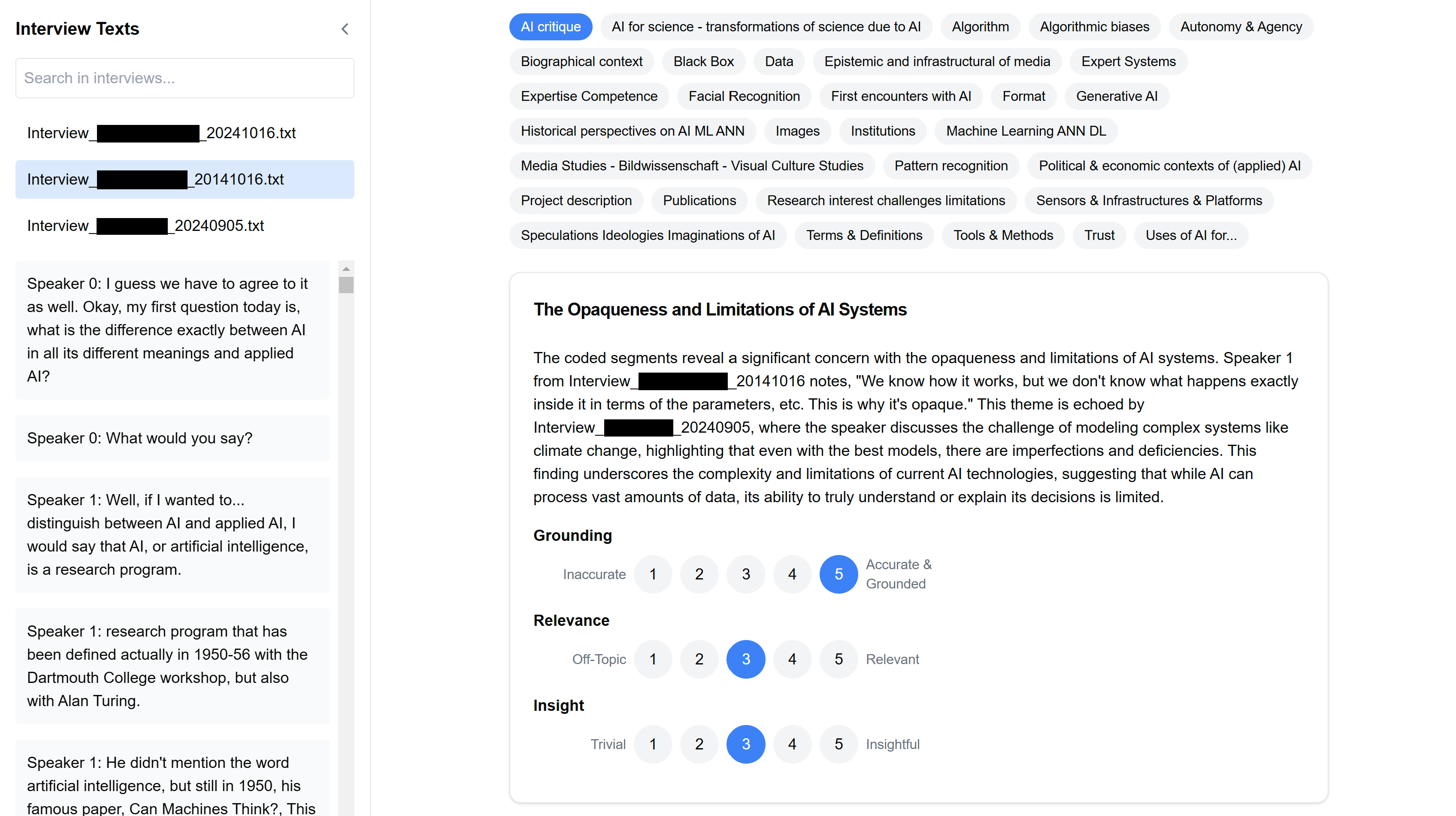}
    \caption{Evaluation interface used by human annotators to assess theoretical findings generated by AICoE}
    \label{fig:findings-evaluation}
\end{figure}

\FloatBarrier

\end{document}